\documentclass[preprint,12pt]{elsarticle}

\usepackage{microtype}
\usepackage{graphicx}
\usepackage{subcaption}
\usepackage{booktabs} 
\usepackage{multirow}
\usepackage[table]{xcolor}
\usepackage{hyperref}

\usepackage{amsmath}
\usepackage{amssymb}
\usepackage{mathtools} 
\usepackage{amsthm}
\journal{Journal of Computational Physics}

\usepackage[capitalize,noabbrev]{cleveref}

\theoremstyle{plain}

\theoremstyle{definition}

\theoremstyle{remark}

\usepackage{cleveref}
\usepackage[textsize=tiny]{todonotes}

\begin{document}
\begin{frontmatter}

  \title{Physics-Informed Neural Networks with Architectural Physics Embedding for Large-Scale Wave Field Reconstruction}
\author[wisc]{Huiwen Zhang}
\ead{hzhang2279@wisc.edu}
\author[wisc]{Feng Ye}
\ead{feng.ye@wisc.edu}
\author[wisc]{Chu Ma\corref{cor}}
\ead{chu.ma@wisc.edu}

\cortext[cor]{Corresponding author}

\affiliation[wisc]{organization={Department of Electrical and Computer Engineering, University of Wisconsin-Madison},
            city={Madison},
            state={WI},
            country={United States}}

\begin{abstract}

Large-scale wave field reconstruction requires precise solutions but faces challenges with computational efficiency and accuracy. The physics-based numerical methods like Finite Element Method (FEM) provide high accuracy but struggle with large-scale or high-frequency problems due to prohibitive computational costs. Pure data-driven approaches excel in speed but often lack sufficient labeled data for complex scenarios. Physics-informed neural networks (PINNs) integrate physical principles into machine learning models, offering a promising solution by bridging these gaps. However, standard PINNs embed physical principles only in loss functions, leading to slow convergence, optimization instability, and spectral bias, limiting their ability for large-scale wave field reconstruction. This work introduces architectural physics embedded (PE)-PINN, which integrates additional physical guidance directly into the neural network architecture beyond Helmholtz equations and boundary conditions in loss functions. Specifically, a new envelope transformation layer is designed to mitigate spectral bias with kernels parameterized by source properties, material interfaces, and wave physics. Experiments demonstrate that PE-PINN achieves more than 10 times speedup in convergence compared to standard PINNs and several orders of magnitude reduction in memory usage compared to FEM. This breakthrough enables high-fidelity modeling for large-scale 2D/3D electromagnetic wave reconstruction involving reflections, refractions, and diffractions in room-scale domains, readily applicable to wireless communications, sensing, room acoustics, and other fields requiring large-scale wave field analysis.
\end{abstract}
\end{frontmatter}







\section{Introduction}

Wave field reconstruction is fundamental across many domains, including precise electromagnetic (EM) field reconstruction in wireless communication and sensing, medical imaging (e.g., ultrasound, thermoacoustic/photoacoustic, and microwave imaging), and non-destructive testing and structural health monitoring. Conventional physics-based numerical methods, such as the Finite Element Method (FEM)~\cite{jin2015finite} and Finite-Difference Time-Domain (FDTD)~\cite{kunz1993finite}, are widely used in high-fidelity simulators (e.g., CST, COMSOL Multiphysics, and ANSYS HFSS). However, these methods require fine spatial discretization (typically $\geq$10 elements per wavelength) to control numerical dispersion, leading to prohibitive memory and computational costs for large-scale or high-frequency problems. As a result, room-scale or larger wave field reconstruction becomes impractical. To address this limitation, alternative approaches such as ray-tracing~\cite{glassner1989introduction} have been developed and are extensively used in EM~\cite{he2018design,german2001wireless} and acoustic modeling~\cite{dushaw1998ray,ameer2010localization,pompei2009computer} for large-scale environments. Based on high-frequency geometric optics approximations, ray-tracing is accurate when the wavelength is much smaller than environmental features (e.g., mmWave/THz bands or frequencies above 5000~Hz in underwater acoustics), but it becomes unreliable when the wavelength is comparable to or larger than the feature size.
Advances in machine learning have spurred data-driven approaches, particularly deep learning, to circumvent the computational cost in wave field modeling in EM~\cite{
xiong2019data,zhang2024hybrid,zhang2020novel,guo2023physics,zhao2023nerf2,zhang2025rf3dgswirelesschannelmodeling}, acoustics~\cite{kexin2022physics,xie2023data,van2025deep}, and other wave domains~\cite{yue2020scalable,tsubaki2020quantum,cheng2024robust}. However, obtaining sufficient training dataset for these approaches in large scale and dynamic environment poses a critical challenge. So far, these models are mostly applied to simplified environments in small scale or near field.
Physics-Informed Neural Networks (PINNs) have recently emerged as a promising alternative for solving problems with known physics constraints~\cite{RAISSI2019686,rasht2022physics,liu2024phase,baldan2023physics,song2022versatile}. By embedding governing partial differential equations (PDEs) and boundary conditions into the loss function, PINNs enable learning from sparse or unlabeled data while maintaining physical consistency. Moreover, unlike traditional mesh-based methods (e.g., FEM and FDTD), PINNs operate on a mesh-free continuous domain, hence significantly mitigating the need for large memory. However, despite the promise, standard PINNs exhibit fundamental limitations in high-frequency electromagnetic wave reconstruction. The primary barrier arises from the inherent spectral bias of neural networks, which prioritize learning low-frequency components~\cite{rahaman2019spectral} while systematically underrepresenting the rapid oscillations of high-wavenumber fields. Compounding this issue, PINNs frequently encounter severe optimization instabilities when modeling singular sources\cite{10.1093/gji/ggab010} or sharp material discontinuities\cite{jagtap2020extended}, phenomena that are ubiquitous in real-world wave physics problems, including wireless systems (e.g., antenna arrays, reconfigurable intelligent surfaces, and complex indoor channels), acoustic propagation in heterogeneous media, and wave interactions with complex engineered or natural structures. Moreover, existing PINNs require significant training time in solving 3D problems, even in a small scale domain that spans a couple of wavelengths~\cite{schoder2024feasibility}. Consequently, standard PINNs cannot reliably reconstruct accurate wave fields for practical scenarios demanding spatial fidelity at room-scale or beyond.

To address these limitations and enable scalable wave field reconstruction, this work proposes architectural physics embedded (PE)-PINN, which allows to embed physics guidance directly into the neural network architecture as fundamental structural components, beyond mere constraints in the loss function in standard PINNs. Electromagnetic (EM) wave reconstruction is used in this work as a representative wave physics problem, although the proposed methodology is broadly applicable to other wave phenomena. In PE-PINN, we consider solutions of the Helmholtz equation governing the electric field $E_z$. To overcome the spectral bias inherent in standard PINNs when representing highly oscillatory wave solutions, we introduce a physics-guided envelope transformation layer that maps the original wave field into a spatially smoother representation.
Specifically, kernels in this layer employ: (i) direction/wavenumber-adaptive plane waves; and (2) center-location/wavenumber-adaptive spherical waves. By parameterizing kernels using physical properties (e.g., source positions, material discontinuities, and Snell's Law that governs wave reflection and diffraction~\cite{peatross2015physics}), the layer analytically factorizes the electric field into learnable envelopes and oscillatory carriers. This demodulates high-frequency components, shifting the neural network's optimization objective from rapid oscillations to smooth envelope functions. The result is accelerated convergence that reduces training iterations by orders of magnitude while maintaining fidelity. Complementing this, we introduce a source-aware hybrid architecture featuring: (i) incident and scattered field separation for scattering problems; (ii) material-aware domain decomposition for heterogeneous media; and (iii) Residual-Based Adaptive Refinement (RAR) with dynamic pruning that concentrates sampling points in high-residual regions. An optional adaptive weighting scheme further stabilizes convergence under competing physical constraints. Extensive 2D/3D numerical experiments confirm that PE-PINN achieves significantly faster convergence and superior reconstruction accuracy compared to baseline PINN methods. Crucially, while conventional PINNs fail to converge to the reference solution within practical training durations (tested over 26 hours), PE-PINN attains convergence within 18 minutes. When validated against FEM modeling in COMSOL Multiphysics under identical physics fidelity, PE-PINN reduces memory requirements by orders of magnitude, particularly in large-scale 3D scenarios. For specific problems studied, COMSOL would require 12.5 TB of theoretical maximum RAM (CPU-based FEM solver storing the sparse stiffness matrix) to achieve equivalent fidelity, whereas PE-PINN achieves comparable results using less than 24 GB of GPU VRAM.

The main contributions of this work are summarized as follows: (1) We propose PE-PINN, which embeds physics into the neural network structure, moving beyond merely incorporating physics in the loss-function in standard PINNs; (2) We derive a multi-component kernel-envelope representation of the wave field that explicitly decouples the field into a known high-frequency kernel function and a learnable low-frequency envelope. The number and formulations of the kernel functions are designed based on prior physical knowledge. This spatial-frequency decoupling mitigates spectral bias and enables unprecedented accuracy in wave field reconstruction in large-scale environments (e.g., with a spatial dimension of more than 20 wavelengths), where standard PINNs fail to converge; (3) We implement material-aware domain decomposition and incident and scattered wave separation to handle complex environments. This method automatically stitches sub-networks across heterogeneous media, overcoming the scalability limits of prior methods in large-scale wave reconstruction with reflections and refractions; and (4) Extensive evaluations are conducted to demonstrate that PE-PINN achieves unprecedented efficiency and accuracy in solving previously intractable problems, including large-scale 2D/3D EM wave reconstruction with reflections, refractions, and diffraction in meter-scale domains. It outperforms standard PINNs in fidelity and achieves at least one order of magnitude speedup. It also demonstrates an orders-of-magnitude reduction in memory requirements compared to COMSOL Multiphysics.


{\color{blue}


}

\section{Wave Physics in PE-PINN}\label{sec:mathematical_formulation}

In conventional PINN-based wave field modeling, the Helmholtz equation and boundary conditions are enforced through mean squared error (MSE) residuals in the training loss~\cite{10.1093/gji/ggab010}. However, relying solely on residual-based optimization often leads to slow convergence and limited accuracy, particularly for high-frequency and large-scale problems. To address these challenges, we move beyond physics enforcement through the loss function alone and incorporate physics-guided design principles directly into key components of the PE-PINN architecture. This integration enables more efficient optimization, improved scalability, and enhanced reconstruction accuracy. Table~\ref{tab:notation} summarizes the notations used throughout this work.

\begin{table}[ht!]
\caption{Summary of notation and symbols.}
\small
\label{tab:notation}
\vspace{-3mm}
\begin{center}
\begin{tabular}{ll}
\toprule
\textbf{Symbol} & \textbf{Description} \\
\midrule
$\mathbf{x}$ & Spatial coordinates in domain $\Omega$ \\
\hline
$E_z^{\{\cdot\}}$ & Complex electric field \\
\hline
$k$ & Wavenumber\\

\hline
$\epsilon, \mu$ & Permittivity, permeability \\
\hline
$\kappa$ & Dielectric constant \\
\hline
$\mathcal{E}_m, M$ & $m$-th component, total \#components \\
\hline
$A_m, \Psi_m, w_m$ & Envelope, Kernel, spatial gate \\
\hline
$\mathbf{d}_m$ & Wave direction parameters \\
\hline
$\mathcal{N}(\cdot; \Theta)$ & Neural network with parameters $\Theta$ \\
\hline
$\Omega, \Gamma$ & Domain and boundaries/interfaces \\
\hline
$\mathcal{L}_{\{\cdot\}}$ & Loss terms\\
\hline
$\lambda_i$ & Loss weights \\

\bottomrule
\end{tabular}
\end{center}
\vspace{-3mm}
\end{table}

\subsection{Wave Physics in PE-PINN Loss Function}

\subsubsection{Helmholtz equation}

The Helmholtz equation is a fundamental PDE that describes time‑harmonic wave phenomena in physics. In the frequency domain, under the assumption of sinusoidal time dependence, the Helmholtz equation is written as $\nabla^2\mathbf{u} + k^2 \mathbf{u} = 0$, relating spatial variations of a field $\mathbf{u}$ to its wavenumber $k$. The equation models propagation, reflection, diffraction, refraction, and other wave behaviors in EM, acoustics, and other wave physics domains. Solutions of the Helmholtz equation depend on source properties and environmental properties, such as material properties, boundary conditions, and geometry. Focusing on the electrical field in the $z$ direction, the Helmholtz equation is:
\begin{equation}
\nabla^2 E_z(\mathbf{x}) + k^2 E_z(\mathbf{x}) = 0,
\end{equation}
where $\nabla^2 $ is the Laplacian operator. The spatial coordinate vector is denoted by $\mathbf{x}$, which is defined as $\mathbf{x}=\{x,y\}$ in the 2D domain and $\mathbf{x}=\{x,y,z\}$ in the 3D domain. $E_z(\mathbf{x}) = E_z^{\text{re}}(\mathbf{x}) + j E_z^{\text{im}}(\mathbf{x})$ is a complex solution. Note that the source excitation is not included as an explicit forcing term; instead, it is enforced through a prescribed field distribution on the source boundary, as detailed in the source excitation loss $\mathcal{L}_\text{src}$ (Section~3.3).

\subsubsection{Boundary conditions}

To ensure the physical validity and well-posedness of the solution, distinct boundary conditions are enforced depending on the boundary type, including Absorbing Boundary Condition for outer truncation, Perfect Electric Conductor (PEC) condition for conductors, and Interface conditions for material continuity. Throughout this work, we consider non-magnetic media with $\mu = \mu_0$, so that material contrast enters only through the permittivity $\epsilon$.


\noindent \textbf{Absorbing Boundary Condition (Outer Truncation):}
To simulate an unbounded medium within a truncated computational domain, the absorbing boundary condition is applied on the outer boundaries $\Gamma_{\text{ext}}$. We employ the First-order Sommerfeld radiation condition to eliminate reflections, which requires the field to satisfy:
\begin{equation}
    \partial_n E_{\Gamma_{\text{ext}}} - i k E_{\Gamma_{\text{ext}}} = 0,
    \label{eq:bc1}
\end{equation}
where $k$ is the wavenumber and $\partial_n$ denotes the derivative along the outward unit normal.

\noindent \textbf{PEC Boundary Condition (Zero Total Field): }
For scenarios involving PECs, the tangential electric field must vanish on the conductor surface $\Gamma_{\text{PEC}}$. This physical constraint mandates that the total electric field satisfies:
\begin{equation}
    E_{\Gamma_{\text{PEC}}} = 0.
    \label{eq:bc2}
\end{equation}
\noindent \textbf{Interface Continuity Condition (Multi-Material Coupling): }
In heterogeneous media, the domain is partitioned into sub-regions to handle varying material properties. At an interface $\Gamma_{\text{int}}$ separating two domains (denoted $\Omega^-$ and $\Omega^+$), electromagnetic theory mandates the continuity of both the electric field and its normal derivative. These continuity conditions are expressed as:
\begin{equation}
\begin{aligned}
    E^{-}_{\Gamma_{\text{int}}} - E^{+}_{\Gamma_{\text{int}}} &= 0, \text{~and~}   \partial_n E^{-}_{\Gamma_{\text{int}}} - \partial_n E^{+}_{\Gamma_{\text{int}}} = 0.
    \label{eq:bc3}
\end{aligned}
\end{equation}

\subsection{Wave Physics in PE-PINN Architecture}

\subsubsection{Envelop Transformation based on Wave Behavior}

We approximate the wave field spatial distribution as a sum of $M$ components, each represented by the product of a pre-assigned oscillatory kernel $\Psi_m(\mathbf{x})$ and a learnable envelope $A_m(\mathbf{x})$:
\begin{equation}
E_z(\mathbf{x})=\sum_{m=1}^{M} w_m(\mathbf{x})A_m(\mathbf{x})\Psi_m(\mathbf{x})=\sum_{m=1}^{M}\mathcal{E}_m,
\label{eq:field_superposition}
\end{equation}
where $m=1,\dots,M$. Both $\Psi_m(\mathbf{x})$ and $A_m(\mathbf{x})$ are complex with both real and imaginary parts. The term $w_m(\mathbf{x}) \in (0, 1)$ is a learnable spatial weight that adaptively regulates the contribution of each wave component across the domain.

The kernel functions $\Psi_m(\mathbf{x})$ are oscillatory carriers capturing the rapidly varying phases of the wave field, so that the network primarily learns the slowing varying envelope $A_m(\mathbf{x})$. The expressions of $\Psi_m(\mathbf{x})$ are predetermined by the physical laws that govern wave behaviors in the modeling domain, such as propagation, reflection, diffraction, and refraction. For \emph{plane waves}, $\Psi_m(\mathbf{x})=e^{-jk\mathbf{d}_m^{\top}\mathbf{x}}$, where $\mathbf{d}_m$ is a unit direction vector satisfying $\|\mathbf{d}_m\|_2=1$. Different vectors $d_m$ represent different propagation directions of the plane waves. For \emph{spherical waves}, $\Psi_m(\mathbf{x})=e^{-jk\|\mathbf{x}-\mathbf{x}_m\|_2}$, where $x_m$ is the center location of the spherical wave.



When plane waves encounters the interface of two materials, reflection and refraction happen. The directions $d_m$ of the reflected and transmitted waves are predetermined by Snell's Law described below. 
    

\vspace{+2mm}
\hrule
\vspace{+2mm}

\noindent \textbf{Snell's Law: }
Let $d_{i,\perp} = \mathbf{d}_i \cdot \mathbf{n}, d_{i,\parallel} = \mathbf{d}_i - d_{i,\perp}\mathbf{n}.$ Then Snell's law requires: $n_1 \lVert d_{i,\parallel} \rVert = n_2 \lVert d_{t,\parallel} \rVert.$ The transmitted direction vector is: $\mathbf{d}_t=\frac{n_1}{n_2} d_{i,\parallel}-\mathbf{n}\sqrt{1 - \left( \frac{n_1}{n_2} \right)^2\lVert d_{i,\parallel} \rVert^2}$.   
\vspace{+2mm}

\hrule

\vspace{+2mm}


When a spherical wave encounters the interface between two materials, the reflected wave is also spherical, with its center located at the mirror point of the original source with respect to the interface. Note that the transmitted wave through the interface is no longer a single spherical wave. Instead, it is the superposition of spherical waves with the centers along a line. The kernel function in this case is still a single spherical wave, with the center location at a mid-point along this line. This spherical kernel can effectively slow down the envelope oscillation. 

The spatial gating factor $w_m$ in Eq.~\eqref{eq:field_superposition} limits the incident, reflected, and transmitted waves to their corresponding side of the material interface.

\subsubsection{Separation of Incident Field and Scattered Field}
The total electric field $E_z^{\text{tot}}$ can be viewed as the superposition of an incident field $E_z^{\text{inc}}$ (generated by the source) and a scattered field $E_z^{\text{sct}}$ (generated by the medium interaction):
\begin{equation}
E_z^{\text{tot}}(\mathbf{x}) = E_z^{\text{inc}}(\mathbf{x}) + E_z^{\text{sct}}(\mathbf{x}),
\label{eq:decomposition}
\end{equation}
where $E_z^{\text{sct}}$ satisfies the radiation condition at infinity. Based on the separation, the implemented PE-PINN is composed of two sub neural networks coupled together by the loss function. The two sub networks are trained to output $E_z^{\text{inc}}(\mathbf{x})$ and $E_z^{\text{sct}}(\mathbf{x})$, respectively. .

\subsubsection{Material-aware domain decomposition}

The environment where EM wave propagates usually contain materials with different electrical and magnetic properties, e.g., permittivity ($\epsilon$), permeability ($\mu$), conductivity ($\sigma$), etc. In a typical room environment, the dielectric constant ($\kappa=\epsilon/\epsilon_0$, also called effective permittivity) could take values of 2-3 for wood/plastic, 4-10 for concrete/brick/glass/ceramic, 80 for water, and infinity for metal. Different material properties lead to different refractive index ($n$) and wavenumber ($k$), following $n=\sqrt{\kappa\mu/\mu_0}$ and $k=2\pi f\sqrt{\epsilon\mu}$, which determines wave behavior inside it. We will decompose regions with different materials into sub-domains and model each sub-domain with a separate neural network. This will enable us to effectively model the discontinuity at material interfaces and tailor the design of the neural network architecture for each sub-domain.

\vspace{-2mm}

\section{Methodology}\label{sec:methodology}

\begin{figure*}[ht!]
    \centering
    \fbox{\includegraphics[width=\linewidth,trim=90 0 40 0, clip]{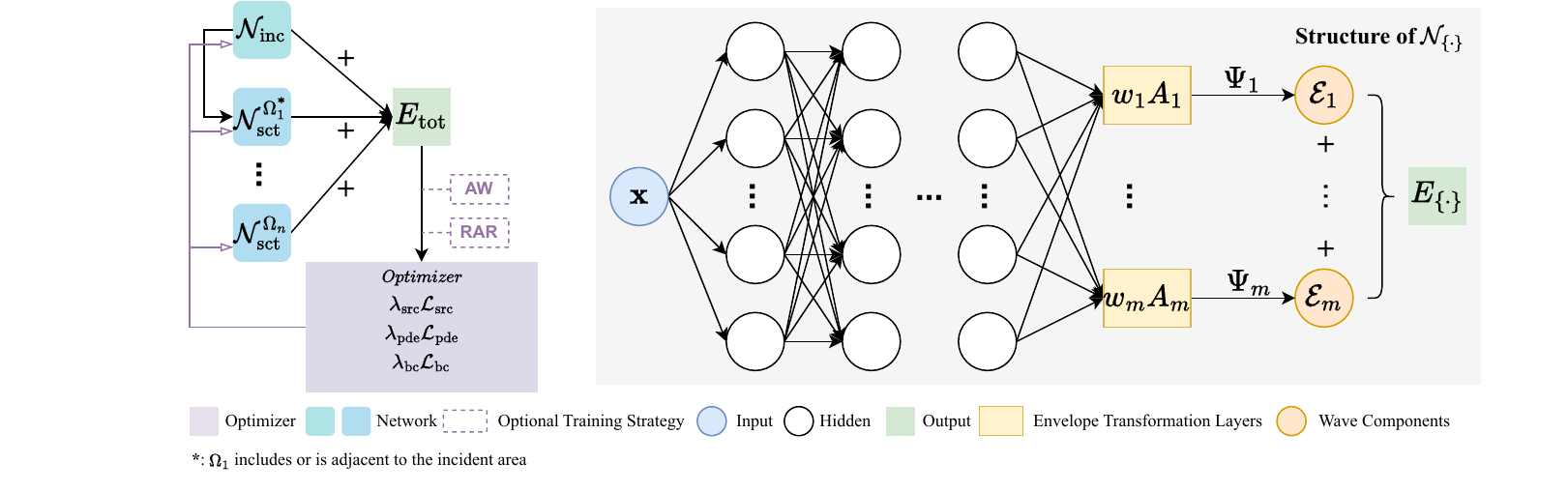}}
     \caption{Overview of the proposed PE-PINN.}
    \label{fig:framework}
\end{figure*}

\subsection{Overview of PE-PINN}

Fig.~\ref{fig:framework} depicts an overview of the proposed PE-PINN. Denoted as $\mathcal{N}(\mathbf{x}; \Theta)$, the PE-PINN framework is to approximate the spatial distribution of the complex electric field $E(\mathbf{x})$. A PE-PINN is composed of multiple sub neural networks (sub-networks) coupled together through the loss function. The main body of each sub-network is constructed as fully-connected layers. We employ periodic sine functions (i.e., $\sigma(x) = \sin(x)$) as the activation functions in PE-PINN. Because sine activations are more effective in mitigating spectral bias and are superior in modeling wave-like physics, enabling the precise calculation of high-order spatial derivatives required by the Helmholtz operator~\cite{rahaman2019spectral, sitzmann2020implicit}.

\subsection{Envelop Transformation Layers}

A distinct contribution of our work is the introduction of envelope transformation layers to enforce physics embedding. As illustrated in Fig.~\ref{fig:framework}, these layers are positioned after the fully connected layers. Unlike conventional PINNs that output the wave field directly, our architecture utilizes the envelope transformation to separate fast and slow variations of the wave field (Eq.~\ref{eq:field_superposition}). The fully connected layers are trained to approximate the slowly varying learnable envelope functions $A_m(\mathbf{x})$. The fast variations are represented by pre-assigned kernel functions ($\Psi_m(\mathbf{x})$) acting as reconstruction heads in the envelope transformation layer that decomposes the field into multiple components. In the implementation, the number of reconstruction heads or kernel functions equals to the number of wave components.
This structural decomposition allows the model to explicitly separate and handle the slow and fast variation behaviors of wave propagation, embedding physical consistency directly into the solution synthesis process rather than relying solely on soft constraints in the loss function.

\subsection{Loss Function}\label{sec:loss_formulation}

A composite loss function $\mathcal{L}$ is formulated to enforce physical laws inside the domain, boundary conditions on the frontiers, and continuity constraints across material interfaces. The total loss is defined as a weighted sum of individual components:
\vspace{-2mm}
\begin{equation}\label{eq:total_loss}
    \mathcal{L}(\Theta) = \lambda_{\text{src}}\mathcal{L}_{\text{src}} + \lambda_{\text{pde}}\mathcal{L}_{\text{pde}} + \lambda_{\text{bc}}\mathcal{L}_{\text{bc}},
    \vspace{-3mm}
\end{equation}

where $\lambda_{\{\cdot\}}$ are weighting coefficients that balance the contribution of each term which can be fixed or adaptive.

$\mathcal{L}_{\text{src}}$ defines the \textbf{Source Excitation Loss}, which guarantees the accurate generation of wave excitation. In our models, the excitation was implemented by assigning a pressure distribution at designated spatial locations (e.g., a Gaussian distribution along a line for plane wave excitation and a constant along a circle for spherical wave excitation). $\mathcal{L}_{\text{src}}$ is formulated as the mean square error between the model output and the assigned distribution.


$\mathcal{L}_{\text{pde}}$ defines the \textbf{PDE Loss}, which serves as the fundamental physical constraint for the neural network. It enforces the Helmholtz equation at $N_{\text{pde}}$ points sampled within the computational domain $\Omega$:
\begin{equation}
    \mathcal{L}_{\text{pde}} = \frac{1}{N_{\text{pde}}} \sum_{i=1}^{N_{\text{pde}}} \left\| \nabla^2 \hat{E}(\mathbf{x}_i) + k(\mathbf{x}_i)^2 \hat{E}(\mathbf{x}_i) \right\|^2,
\end{equation}
 where $\hat{E}$ denotes the field variable being solved (representing the incident field $E_{\text{inc}}$ or scattered field $E_{\text{sct}}$, and $k(\mathbf{x}_i)$ represents the local wavenumber.

$\mathcal{L}_{\text{bc}}$ defines the \textbf{Boundary Loss ($\mathcal{L}_{\text{bc}}$)}, which implements the boundary conditions (Eqs.~\eqref{eq:bc1}, \eqref{eq:bc2}, and \eqref{eq:bc3}), including scattering boundaries, perfect electrical conductors, and material interfaces, as mean square errors.

\subsection{Sub-nets for Separating the Incident and Scattered Fields}


In PE-PINN, we employ the separation of the incident and scattered fields (Eq.~\ref{eq:decomposition}) for modeling scenarios that have refrlection, refraction, or diffraction caused by scatterers/discontinuities. In these scenarios, dedicated neural networks, $\mathcal{N}_{\text{inc}}(\mathbf{x}; \Theta_{\text{inc}})$ and $\mathcal{N}_{\text{sct}}(\mathbf{x}; \Theta_{\text{sct}})$, are utilized to model the incident and scattered fields, respectively. For free-space propagation, only the incident network $\mathcal{N}_{\text{inc}}(\mathbf{x}; \Theta_{\text{inc}})$ is employed. When field separation is employed, both the $\mathcal{N}_{\text{inc}}(\mathbf{x}; \Theta_{\text{inc}})$ and $\mathcal{N}_{\text{sct}}(\mathbf{x}; \Theta_{\text{sct}})$ are trained simultaneously using a single global optimizer, ensuring synchronized convergence. 
It is important to note that the two networks optimize distinct objective functions: $\mathcal{N}_{\text{inc}}$ is updated exclusively based on the source loss $\mathcal{L}_{src}$ and its intrinsic PDE loss $\mathcal{L}_{pde}^{inc}$. Meanwhile, $\mathcal{N}_{\text{sct}}$ minimizes a separate scattered PDE loss $\mathcal{L}_{pde}^{sct}$ and boundary condition loss $\mathcal{L}_{bc}$, where the output of $\mathcal{N}_{\text{inc}}$ serves as the driving term. A gradient detachment operation is implemented to ensure that the optimization of the incident field remains independent of the scattered field. 


\subsection{Sub-nets for Spatial Domains with Different Material Properties}
Furthermore, to model wave propagation through heterogeneous media with varying electromagnetic properties, a material-aware domain decomposition strategy is adopted~\cite{JAGTAP2020113028}. The global domain $\Omega$ is partitioned into non-overlapping sub-domains $\Omega_i$, such that each sub-domain represents a homogeneous medium with a constant local wavenumber $k_i$.  A distinct neural network is assigned to each sub-domain. To ensure a physically consistent global solution, these independent sub-networks are mathematically stitched together at the material interfaces by enforcing continuity constraints (Eq.~\ref{eq:bc3}) directly in the loss function, the detailed formulation of which is presented in Section \ref{sec:loss_formulation}. This decomposition mitigates the spectral bias often encountered in single-network PINNs when approximating functions with sharp gradient discontinuities across material interfaces.

\subsection{Training Strategy} 

One notable issue in wave field reconstruction is that the error distribution is often non-uniform, with high residuals concentrated near obstacles or regions with complex interference patterns. Using a fixed set of points may fail to capture these high-frequency features. To address this, we employ a modified Residual-Based Adaptive Refinement (RAR) method \cite{lu2021deepxde}. Unlike the standard implementation, which monotonically increases the dataset size, our approach incorporates a dynamic pruning mechanism to maintain computational efficiency, described as follows: A dense pool of candidate points is first sampled. Points with the highest residuals (top-$k$) are identified as ``hard examples'' and appended to the points set $N_{\text{pde}}$. To manage computational resources, we define a maximum dataset capacity $N_{\text{max}}$. The point set consists of a fixed base set $N_{\text{pde}}^{\text{base}}$ and a dynamic adaptive set $N_{\text{pde}}^{\text{adapt}}$. When new hard examples are identified and the total capacity is reached, we employ a selective replacement strategy: the new points replace the points with the lowest residuals exclusively from the previously accumulated adaptive set $N_{\text{pde}}^{\text{adapt}}$. This mechanism ensures that the initial uniform coverage is preserved, while the adaptive buffer continuously evolves to discard well-learned regions. This refinement process is performed periodically during training. The revised RAR strategy ensures high-fidelity reconstruction in critical regions without suffering the overhead of an ever-growing dataset. (See ~\ref{appendix B} for more on  training strategies.)

\subsection{Kernel Selection Guidelines}
\label{sec:kernel_selection}

A distinctive feature of PE-PINN is that the kernel functions $\Psi_m(\mathbf{x})$ in the envelope transformation layer are not free parameters but are predetermined from the known physics of the problem. Each kernel corresponds to a dominant wave mechanism identifiable from the problem geometry, and the selection follows a systematic procedure grounded in wave physics.

The source kernel is chosen to match the dominant radiation pattern of the excitation. A point source corresponds to a spherical wave kernel centered at the source location, while a plane wave source corresponds to a plane wave kernel along the propagation direction. When the radiated field cannot be adequately described by a single wave type, multiple kernels can be combined to capture different spatial regions of the field.

When the domain contains reflectors, a kernel is assigned for the reflected wave. Its type is inherited from the incident wave: a plane wave reflecting off a flat surface remains a plane wave with its direction determined by the law of reflection, while a spherical wave produces a reflected spherical wave centered at the image source.

Diffraction at scatterer corners and vertices generates secondary spherical waves emanating from the diffracting point. Each such feature is assigned a spherical wave kernel centered at that location.


The most involved case arises for transmitted waves through a flat interface into a medium with a different refractive index. For a plane wave, the transmitted wave remains a plane wave with a direction given by Snell's law. For a spherical wave, the transmitted field is a continuous superposition whose effective virtual centers lie along the refraction caustic behind the interface, and no single-center spherical kernel matches it exactly.  PE-PINN places $\mathbf{x}_m$ on this caustic using optical-path-length matching at a representative angle within the domain: the paraxial limit recovers the geometric image $x_{\text{int}} - (n_2/n_1)\,d$, while matching at larger angles yields virtual centers further behind the interface. The learnable envelope $A_m(\mathbf{x})$ absorbs the residual phase mismatch over the rest of the domain.

\section{Experiments and Evaluation}\label{sec:evaluation}

\subsection{Settings and Studied Scenarios}

We implemented the proposed PE-PINN framework for reconstructing EM wave distribution in a room-sized space ($5\times5~\text{m}^2$ in 2D and $5\times 5\times 5~\text{m}^3$ in 3D). The frequency of the EM wave is 2.4 GHz, corresponding to a wavelength of 12.5 cm. The size of $5~\text{m}$ contains 40 wavelengths. Such fast field oscillation brings a significant challenge in conventional FEM and PINN methods. Table~\ref{tab:scenarios} summarizes the EM field reconstruction scenarios we have implemented. In the ``Environment'' column, the ``Free space'' refers to scenarios modeling free-space wave propagation with no scattering media in the spatial domain. The ``Reflection'' refers to a flat reflector aligned along the vertical direction. The ``Diffraction'' refers to wave diffraction by a square (2D) or cubic (3D) PEC located at the center of the spatial domain. The ``Refract. (H)" refers to a flat material interface with dielectric constant $\kappa=1$ on the left and $\kappa=9$ on the right. The ``Refract. (S)'' refers to a vertical material strip, with $\kappa=9$ and a width of 0.9 m, placed at the center of the $x$ axis, generating two flat vertical material interfaces at $x=-0.45$ m and $x=+0.45$ m, respectively. In the ``Source'' column, the ``Plane'' refers to plane wave source with a predescribed $E_z$ field following a Gaussian distribution along  an edge (2D) or a surface (3D) of the spatial domain, and the ``Spherical'' refers to a spherical wave source with a predescribed constant $E$ field along a circle (2D) or sphere (3D). 

Our implementation is based on PyTorch. All models were trained on a workstation with an Intel i9-13900K CPU, 128 GB RAM, and a single NVIDIA RTX 4090 GPU with 24 GB VRAM.

\begin{table}[t]
    \centering
        \scriptsize             
    \setlength{\tabcolsep}{12pt}
    \caption{Configurations of the studied scenarios.}
    \label{tab:scenarios}
    \begin{tabular}{lcccc}
        \toprule
       No. & Environment & Scene & Dimension (m) & Source\\
        \midrule
        1 & Free space & 2D & $5\times5$ & Plane\\
        2 & Free space & 3D & $1\times1\times 1$ & Plane\\
        3 & Free space & 2D & $5\times5$ & Spherical  \\
        4 & Free space & 3D & $5\times5\times 5$ & Spherical \\
         \midrule
        5 & Reflection & 2D & $5\times5$ & Plane\\
        6 & Reflection & 2D & $5\times5$ & Spherical  \\
         \midrule
        7 & Diffraction & 2D & $5\times5$ & Spherical  \\
        8 & Diffraction & 3D & $5\times5\times 5$ & Spherical \\
         \midrule
        9 & Refract. (H) & 2D & $5\times5$ & Spherical  \\
        10 & Refract. (H) & 3D & $5\times5\times 5$& Spherical  \\
        11 & Refract. (S) & 2D & $5\times5$ & Spherical \\
        12 & Refract. (S) & 3D & $5\times5\times 5$ & Spherical \\
        \bottomrule
    \end{tabular}
\end{table}


\begin{figure*}[ht]
  \centering
  \includegraphics[width=\textwidth]{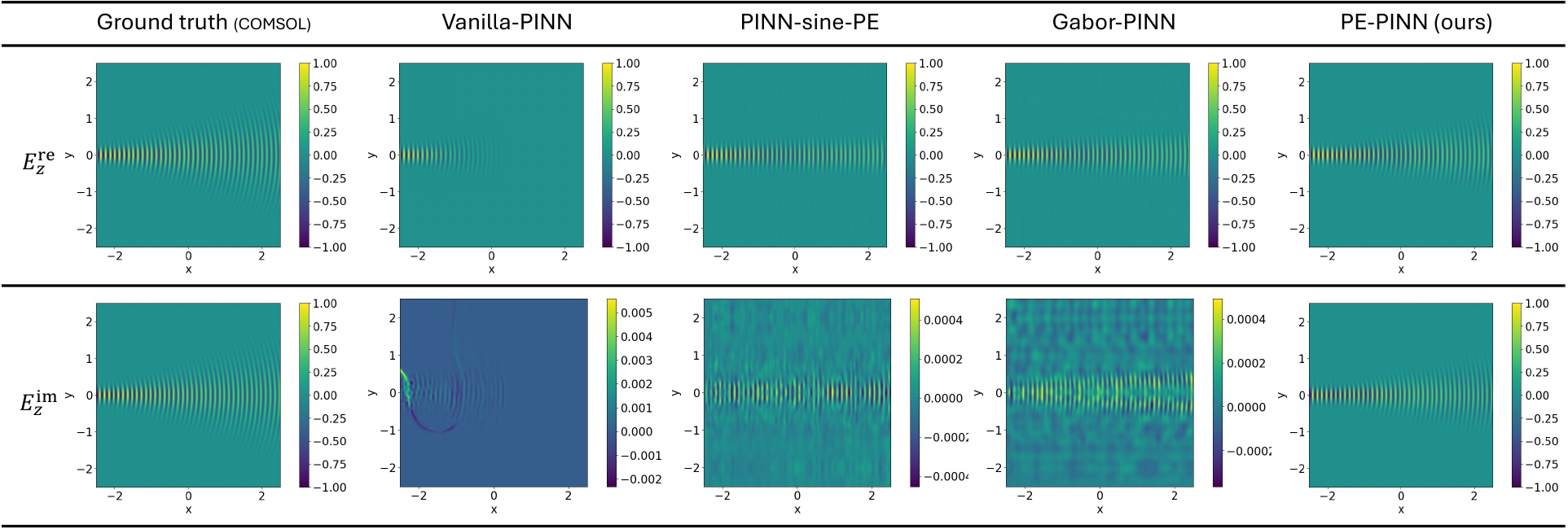}
    \caption{Wave field reconstruction results for 2D free space with plane wave source (Scenario 1).} 
  \label{fig:comparison-results}
\end{figure*}

\begin{table}[ht]
    \centering
    \scriptsize             
    \setlength{\tabcolsep}{4pt}
    \caption{Performance comparison for selected wave field construction scenarios. (See ~\ref{appendix A} for all scenarios).}
    \label{tab:performance_comparison}
    \begin{tabular}{clccccc}
        \toprule
       No.  & PINN Framework & Time/100 Iters & Time (h:m:s) & \#Iters & $\mathcal{L}_{\text{pde}}$ & $\text{MSE}_{\text{(vs.COMSOL)}}^{\text{After\ norm}}$ \textsuperscript{$\dagger$} \\
        \midrule
        \multirow{4}{*}{1} & Vanilla-PINN~\cite{RAISSI2019686} & 0.92 s & 26:27:27 & 10 M & 2.24e-01 & 4.07e-02\\
         & PINN-sine-PE~\cite{inproceedings} & 1.20 s & 06:50:52 & 2 M & 5.24e-02& 2.96e-02\\
         & Gabor-PINN~\cite{abedi2025gaborenhancedphysicsinformedneuralnetworks} & 2.01 s & 02:49:43 & 500 K & 4.78e-02 & 2.79e-02\\
         & PE-PINN & 2.11 s & 00:17:54 & 50 K & 1.11e-01 & 3.38e-03\\
          \midrule
         7 & PE-PINN & 5.11 s & 00:44:57 & 50 K & 7.43e-02 & 5.01e-03 \\
         \midrule
         8 & PE-PINN & 10.02 s & 02:47:06 & 100 K & 1.04e-02 & N/A \\
         \midrule
         9 & PE-PINN & 3.58 s & 01:00:09 & 100 K & 1.36e-00 & 7.12e-03 \\
         \midrule
         10 & PE-PINN & 4.91 s & 02:45:10 & 200 K & 1.55e-01 & N/A \\
        \bottomrule
    \end{tabular}
    \vspace{+2mm}

    \parbox{\linewidth}{\footnotesize
    $\dagger$, $\text{MSE}_\text{(vs.COMSOL)}^\text{After norm}$ denotes the mean squared error between the PE-PINN prediction and the COMSOL reference after normalizing both fields by the source-boundary amplitude.}

\end{table}

\subsection{Benchmarking Existing PINNs for Wave Field Reconstruction}
To validate the effectiveness of PE-PINN, we benchmark it against three established PINN approaches. First, we consider the Vanilla-PINN~\cite{RAISSI2019686}, which serves as the fundamental baseline for physics-informed deep learning. Second, we evaluate a Fourier Feature-based variant denoted as PINN-sine-PE, which incorporates the positional encoding scheme proposed by Tancik et al.~\cite{tancik2020fourierfeaturesletnetworks,inproceedings} to improve high-frequency learning. Finally, we compare our method with Gabor-PINN~\cite{abedi2025gaborenhancedphysicsinformedneuralnetworks}, a recently proposed architecture designed to enhance the high-frequency convergence of the standard PINN-sine-PE baseline. Unlike the physics-embedded designs in our envelope transformation layers, a fixed set of Gabor functions are applied as the kernel functions in Gabor-PINN. The comparison is conducted on a fundamental 2D $5$ m $\times$ $5$ m freespace scenario (Scenario 1 in Table~\ref{tab:performance_comparison}). The spatial domain is specified as $\Omega = [-2.5~\text{m}, 2.5~\text{m}]^2$. An incident source is introduced at the boundary $(-2.5, 0)$ with a Gaussian profile defined as: $E_z^\text{Re}(\mathbf{x}) = \exp\left(- \frac{y^2}{0.2^2}\right)$, and $E_z^\text{Im}(\mathbf{x}) = 0$. 

Note that the field generated by this Gaussian source distribution is not an ideal plane wave. While the wavefront is initially planar at the source boundary ($x = -2.5$), the finite transverse width ($w_0 = 0.2$) introduces diffraction effects during propagation. Consequently, in the far field, the beam will diverge, and the wavefronts will gradually evolve from planar to cylindrical, deviating from the non-diffracting nature of an ideal plane wave. Therefore, PE-PINN decomposes the field into a plane wave component with a kernel function $\Psi_1(\mathbf{x}) = e^{-jkx}$, and a spherical wave component with a kernel function $\Psi_2(\mathbf{x}) = e^{-jk\sqrt{(x+2.5)^2+y^2}}$. Specifically, the network takes spatial coordinates $\textbf{x}$ as input and employs a shared backbone with 4 fully connected layers (widths: 40, 120, 120, 120) using sinusoidal activation functions. 
The outputs from the fully connected layers are then branch into two independent linear layers, accounting for the planar and spherical wave components, respectively. Each envelope transformation kernel projects to 3 output channels ($A_m^{\text{re}}(\mathbf{x})$, $A_m^{\text{im}}(\mathbf{x})$, and $w_m(\mathbf{x})$). Finally, the total electric field $E_z(\mathbf{x})$ are calculated through Eq.~\eqref{eq:field_superposition}.

A comprehensive comparison of computational performance and solution accuracy for this free-space propagation scenario is presented in Fig.~\ref{fig:comparison-results} and Table~\ref{tab:performance_comparison}. Among benchmarking models, the Vanilla-PINN failed to converge after 10 million iterations (over 26 hours); the PINN-sine-PE baseline required 2 million iterations (approximately 7 hours) to converge the real part of the field, but the imaginary part had not yet converged. The Gabor-PINN baseline, while demonstrating improved efficiency with a 60\% reduction in wall-clock time and requiring only one-quarter of the baseline's epochs compared to other PINN baselines, also struggles to converge on the imaginary part of the field. In comparison, the proposed PE-PINN converged in just 50,000 iterations (approx. 18 minutes), effectively speeding up the process by at least one order of magnitude. More importantly, PE-PINN successfully converged on both the real and imaginary parts of the field, demonstrating its superior performance in wave field reconstruction.

\begin{figure}
    \centering
    \includegraphics[width=1\linewidth]{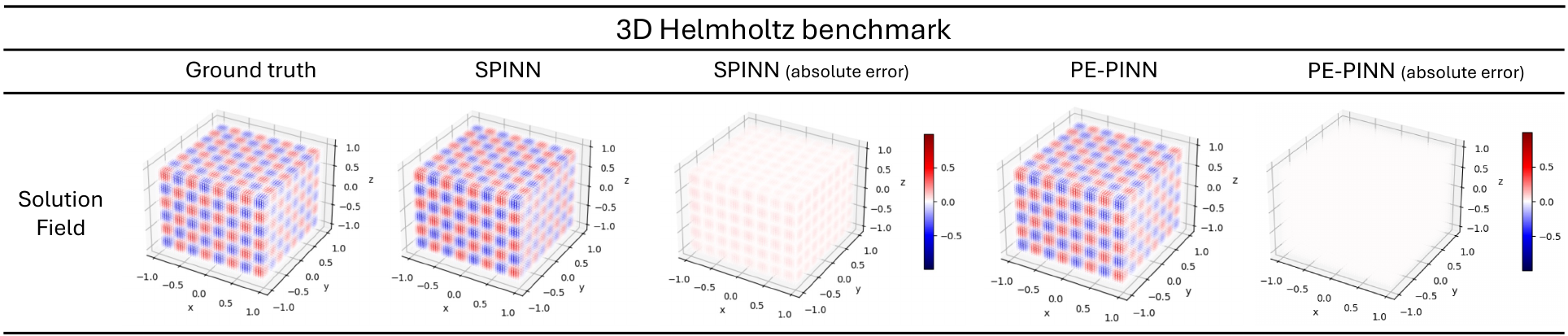}
    \caption{3D Helmholtz benchmark compared with SPINN}
    \label{fig:spinn_compare}
\end{figure}

We further compare PE-PINN with the Separable PINN (SPINN)~\cite{cho2023separablephysicsinformedneuralnetworks}, which addresses the computational challenges of high-dimensional PDEs by representing the solution as a product of low-dimensional components. We evaluate two scenarios. First, on the 3D Helmholtz benchmark from the official SPINN implementation, whose analytical solution is given by $u(x,y,z)=\sin(4\pi x)\sin(4\pi y)\sin(3\pi z)$, PE-PINN converges in 5,000 epochs (61\,s wall-clock) with a relative $L^2$ error of $3.05\times10^{-3}$ and an MSE of $1.11\times10^{-6}$, while SPINN requires 50,000 epochs (50\,s) to reach a relative $L^2$ error of $2.13\times10^{-2}$. Despite using $10\times$ fewer iterations, PE-PINN achieves a relative $L^2$ error nearly an order of magnitude lower, highlighting the advantage of embedding physically meaningful wave structures into the architecture. The visualized comparsion is shown in Fig~\ref{fig:spinn_compare}. Second, on the 2D free-space Gaussian-beam propagation scenario (Scenario~1), SPINN fails to converge to an acceptable solution even with $R=128$ separable components after 5 million iterations, whereas PE-PINN converges in 50K iterations (approx.\ 18 minutes). This is because the Gaussian-beam field does not admit a simple separable factorization in Cartesian coordinates, requiring a prohibitively large number of separable components to approximate the coupled spatial variations.

\begin{figure}[t]
    \centering
    \includegraphics[width=\linewidth]{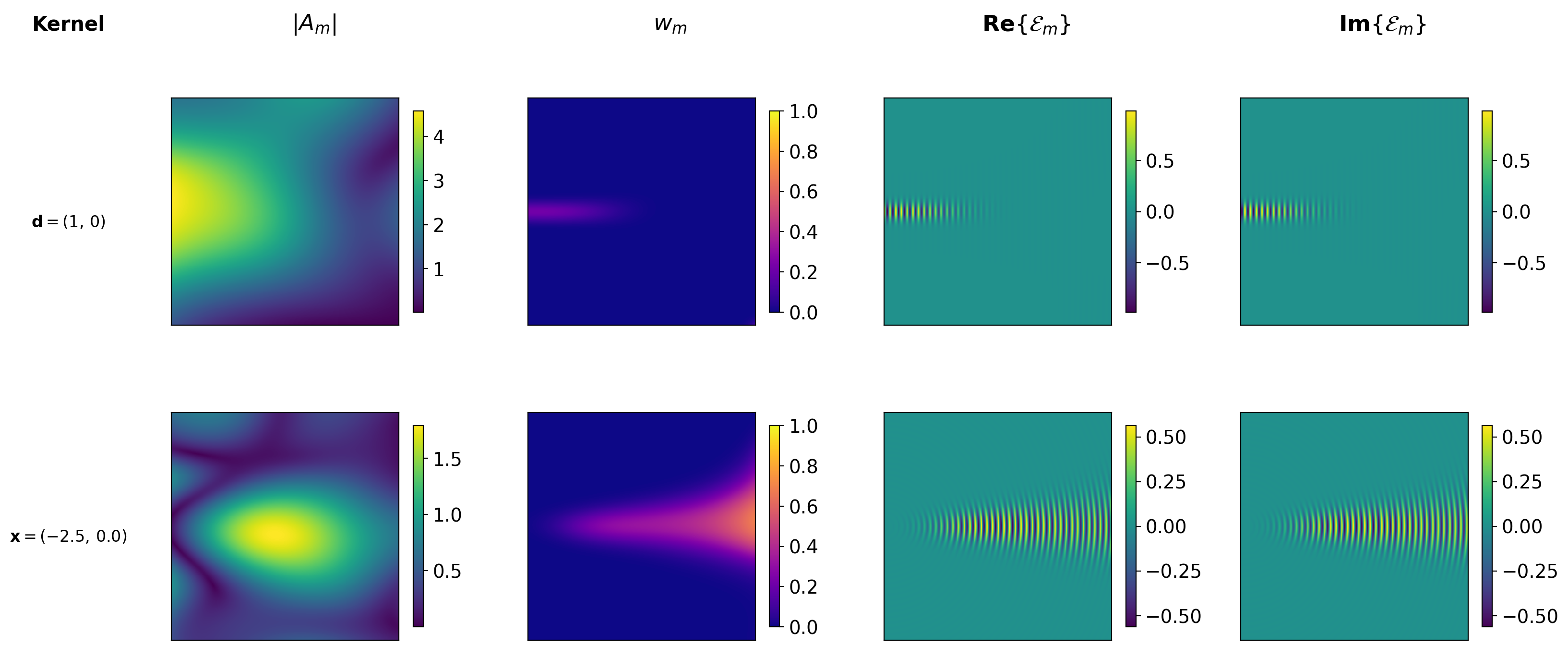}
    \caption{Visualization of the envelope-kernel decomposition for Scenario~1.}
    \label{fig:envelope_freespace}
\end{figure}

To illustrate how the envelope transformation mitigates spectral bias, Fig.~\ref{fig:envelope_freespace} visualizes the learned decomposition for Scenario~1. Each row corresponds to one kernel head. The first column shows the learned envelope magnitude $|A_m(\mathbf{x})|$. The second column shows the learned spatial gate $w_m(\mathbf{x})$, which adaptively regulates the contribution of each component across the domain. The last two columns show the reconstructed wave components $\text{Re}\{\mathcal{E}_m\}$ and $\text{Im}\{\mathcal{E}_m\}$, which recover oscillatory patterns through the product of smooth envelopes and pre-assigned oscillatory kernels. 

\subsection{Results of Wave Field Reconstruction with Diffraction and Refraction}

\begin{figure*}[t]
    \centering
    \includegraphics[width=\linewidth]{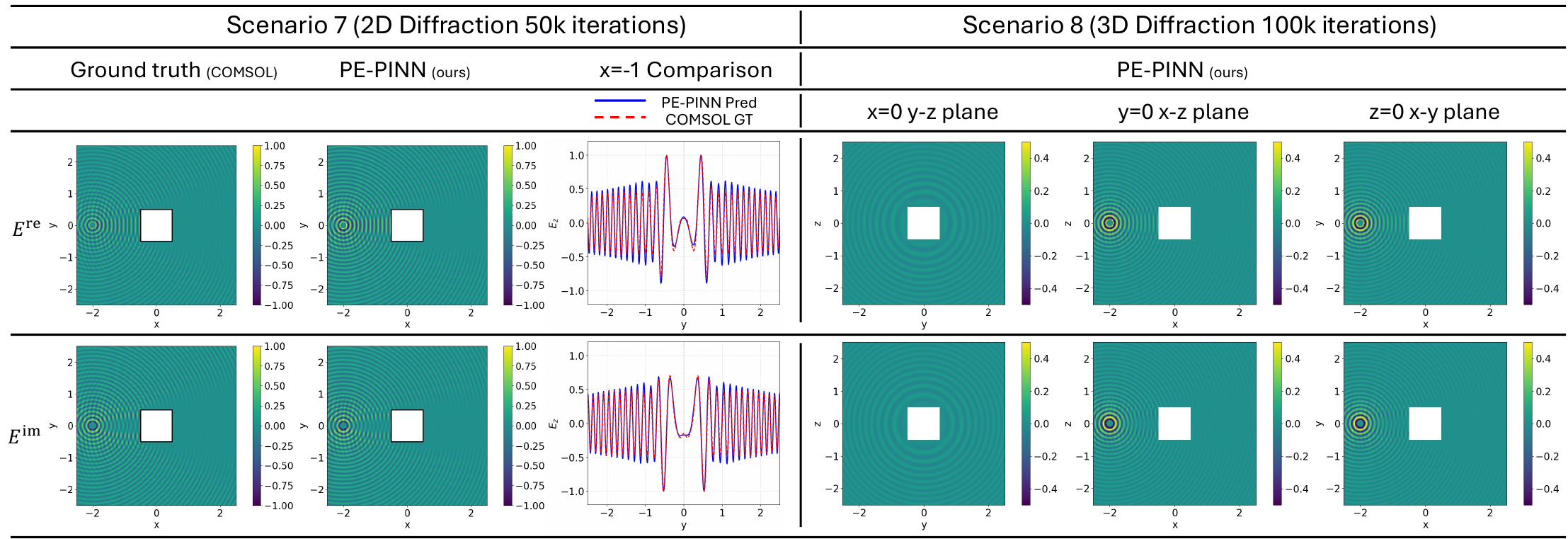}
    \caption{Wave field reconstruction results for 2D  and 3D spherical wave diffraction (Scenario 7, 8).}
    \label{fig:Scenario7&8}
\end{figure*}
\begin{figure*}[t]
    \centering
    \includegraphics[width=\linewidth]{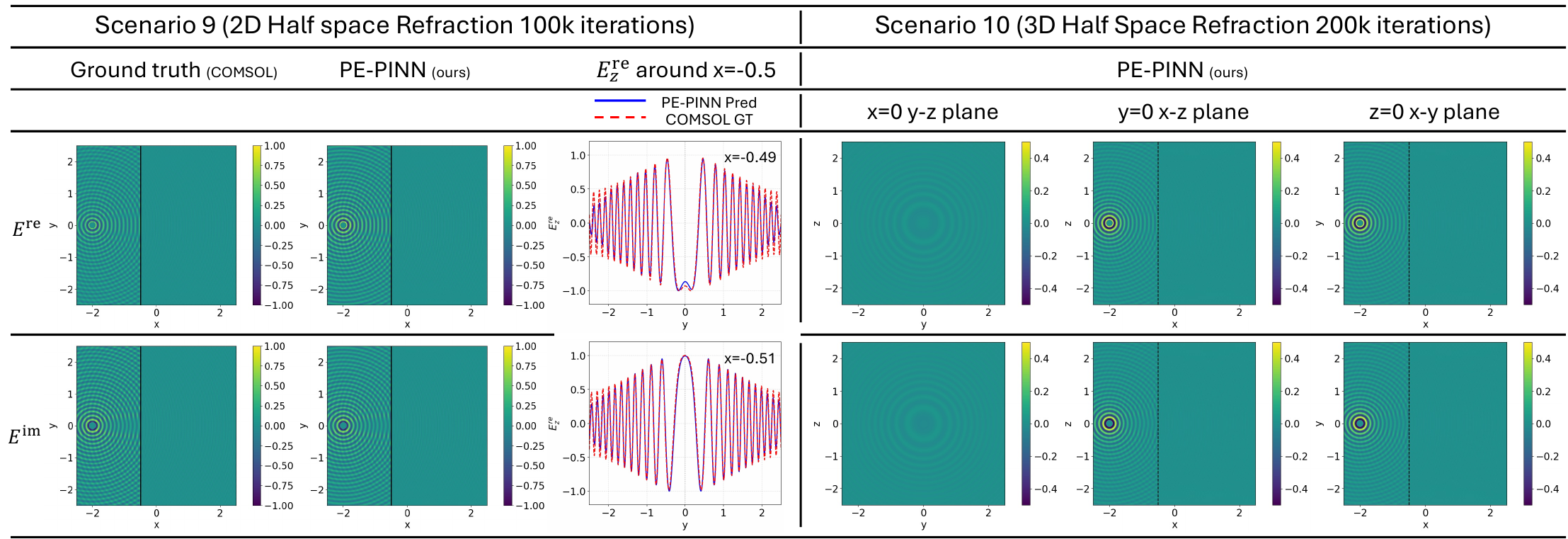}
    \caption{Wave field reconstruction results for 2D and 3D spherical wave half space refraction (Scenario 9, 10).}
    \label{fig:Scenario9&10}
\end{figure*}

\begin{table}[ht]
    \centering

    \caption{Kernel Settings for Scenarios 7, 8, 9, and 10.}
    \label{tab:kernel_selected}
    \begin{tabular}{ccc}
        \toprule
       No.  & \#kernel & Spherical Wave Kernel $ \mathbf{x}_m$  \\
        \midrule
        \multirow{2}{*}{7} & \multirow{2}{*}{4} & $(-2, 0)$, $(1, 0)$, \\
        
                                                     & & $(-0.5, 0.5)$, $(-0.5, -0.5)$ \\
         \midrule
         \multirow{3}{*}{8} & \multirow{3}{*}{6} &  $(-2, 0, 0), (-0.5, -0.5, 0.5)$, \\
                                                     & & $(-0.5, 0.5, 0.5)$, $(1, 0, 0)$\\
                                                     & & $(-0.5, 0.5, -0.5), (-0.5, -0.5, -0.5)$ \\
        \midrule
         9 & 3 &  $(-2, 0)$, $(1, 0)$, $(-7, 0)$ \\
         \midrule
         10 & 3 &  $(-2, 0, 0)$, $(1, 0, 0)$, $(-7, 0, 0)$ \\
        \bottomrule
    \end{tabular}
\end{table}

We then demonstrate the evaluation results of PE-PINN wave field reconstructions in more sophisticated cases with diffraction (Scenarios 7 and 8) and refraction (Scenario 9 and 10). Note that the baseline PINNs can no longer converge within a reasonable amount of time. Therefore, only results from PE-PINN and ground truth FEM solutions obtained from COMSOL Multiphysics (available only in the 2D case) are presented for those selected scenarios. The kernel functions in these scenarios are listed in Table~\ref{tab:kernel_selected}. 
As depicted in Fig.~\ref{fig:Scenario7&8}, PE-PINN successfully reconstructs the wave field with diffraction at high fidelity (MSE at 5.01e-03 from Table~\ref{tab:performance_comparison}) compared with the ground truth, validated by the match between the source and scatterer ($x=-1$), and it shows the reconstructed 3D wave field with diffraction from PE-PINN. Note that the corresponding ground truth is unattainable from COMSOL because the estimated memory requirement is approximately 12.5 TB. 
Further, as depicted in Fig.~\ref{fig:Scenario9&10}, PE-PINN can reach high fidelity (MSE at 7.12e-03 from Table~\ref{tab:performance_comparison}) in wave field reconstruction with refraction. 
Furthermore, it also demonstrates the feasibility of running PE-PINN in 3D room-scale wave field reconstruction with refraction. 

\begin{figure}[htbp!]
    \centering
    \includegraphics[width=\linewidth]{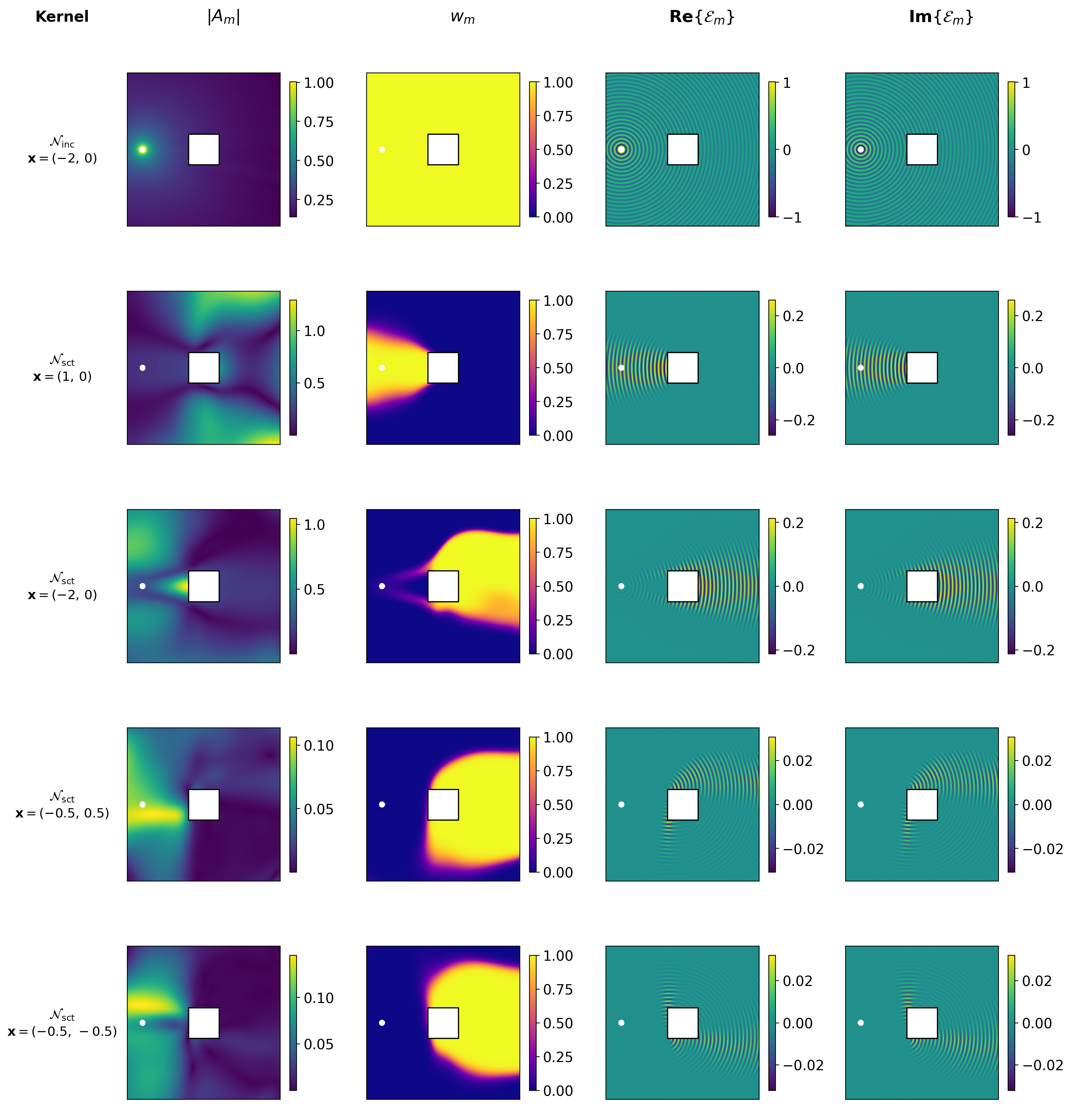}
    \caption{Visualization of the envelope-kernel decomposition for Scenario~7.}
    \label{fig:envelope_diffraction}
\end{figure}

Fig.~\ref{fig:envelope_diffraction} further visualizes the envelope--kernel decomposition for Scenario~7 (diffraction). Unlike the free-space case, the scatter network employs four kernel heads corresponding to distinct physical mechanisms: a reflection kernel centered at the image source location, a shadow compensation kernel at the primary source, and two diffraction kernels at the front corners of the PEC scatterer. The learned envelopes $|A_m|$ remain spatially smooth for all heads, while the spatial gates $w_m$ exhibit physically interpretable patterns---for example, the reflection head concentrates its contribution on the illuminated side of the scatterer, and the diffraction heads activate primarily in the shadow region behind the corners. 

See ~\ref{appendix A} for more results.

\subsection{Ablation Study}
To systematically evaluate the contribution of each proposed component in PE-PINN, we conduct a comprehensive ablation study. Unless otherwise stated, ablation experiments use Scenario~7 as the primary benchmark. Where the analysis involves heterogeneous media, we report results on Scenario~9. We report both the PDE residual loss $\mathcal{L}_\text{pde}$ and the total MSE against the COMSOL reference solution (denoted MSE$_\text{total}$), as the two metrics can behave differently: a lower PDE loss does not always correspond to a lower total MSE, since the PDE residual is only one term in the composite objective and does not account for boundary and interface conditions. All results are summarized in Table~\ref{tab:ablation}.

\begin{table}[htbp]
\centering
        \scriptsize             
    \setlength{\tabcolsep}{12pt}
\caption{Ablation study results. All experiments use Scenario~7 unless marked with $\ast$ (Scenario~9). Bold rows indicate the default configuration for each group.}
\label{tab:ablation}
\begin{tabular}{llcccc}
\hline
Category & Configuration & $\mathcal{L}_\text{pde}$ & MSE$_\text{total}$ & Time (h:m:s) \\
\hline
\multirow{3}{*}{Components}
& \textbf{Full PE-PINN} & \textbf{7.43e-02} & \textbf{5.01e-03} & \textbf{00:44:57} \\
& w/o diffraction heads\textsuperscript{$\dagger$} & 1.02e-01 & 5.44e-03 & 00:29:28 \\
& w/o RAR & 8.87e-02 & 5.11e-03 & 00:44:01 \\
\hline
\multirow{3}{*}{Depth}
& 3 layers & 8.89e-02 & 5.38e-03 & 00:42:43 \\
& \textbf{4 layers} & \textbf{7.43e-02} & \textbf{5.01e-03} & \textbf{00:44:57} \\
& 5 layers & 6.25e-02 & 5.31e-03 & 00:47:43 \\
\hline
\multirow{3}{*}{Width}
& 90 & 1.11e-01 & 5.72e-03 & 00:44:09 \\
& \textbf{120} & \textbf{7.43e-02} & \textbf{5.01e-03} & \textbf{00:44:57} \\
& 150 & 7.34e-02 & 5.45e-03 & 00:47:15 \\
\hline
\multirow{3}{*}{Collocation spacing}
& 0.005\,m ($\sim$25\,pts/$\lambda$) & 9.69e-02 & 5.17e-03 & 00:45:12 \\
& \textbf{0.015\,m ($\sim$8\,pts/$\lambda$)} & \textbf{7.43e-02} & \textbf{5.01e-03} & \textbf{00:44:57} \\
& 0.045\,m ($\sim$3\,pts/$\lambda$) & 9.21e-02 & 5.30e-03 & 00:45:18 \\
\hline
\multirow{2}{*}{Loss weights}
& \textbf{Fixed} & \textbf{7.43e-02} & \textbf{5.01e-03} & 00:44:57 \\
& Adaptive & 8.48e-02 & 6.15e-03 & 00:46:01 \\
\hline
\multirow{2}{*}{Loss weights$^\ast$}
& Fixed & 4.20e-00 & 7.77e-03 & 01:00:04 \\
& \textbf{Adaptive} & \textbf{1.09e-00} & \textbf{6.80e-03} & 01:00:31 \\
\hline

\end{tabular}
\parbox{\linewidth}{\footnotesize
$^\dagger$\,Removes corner diffraction kernels at $(-0.5,0.5)$ and $(-0.5,-0.5)$.}
\vspace{2pt}
\end{table}

Removing the two corner diffraction kernels raises PDE loss by 37\% and total MSE by 9\%, while reducing training time from 44:57 to 29:28 due to the smaller computational graph. This confirms that physics-guided diffraction kernels improve accuracy in the shadow region behind the scatterer. Disabling RAR causes a modest degradation in both metrics at negligible time savings, indicating that adaptive collocation refinement provides a small but consistent benefit.

Increasing depth from 4 to 5 layers lowers the PDE residual but worsens total MSE; reducing to 3 layers degrades both. Widening from 120 to 150 neurons yields comparable PDE loss but worse MSE; narrowing to 90 degrades both. For collocation density, the default spacing of 0.015\,m is optimal; denser sampling adds no benefit while sparser sampling under-resolves the oscillations. Training times remain comparable across all density settings ($\sim$45 minutes), consistent with the observation that PE-PINN's cost doesn't scale with the number of collocation points.

For the loss weighting strategy, the optimal one is scenario-dependent. Fixed weights work better for the single-medium diffraction case, while adaptive gradient-normalized weighting is preferable for the refraction scenario with large dielectric contrast, where it dynamically balances competing gradients from PDE, boundary, and interface losses. Training times are comparable under both strategies, so the choice should be guided by problem complexity.

\section{Discussions}\label{sec:discussion}

PE-PINN has demonstrated successful wave field reconstruction for 2D and 3D wave propagation, reflection, diffraction, and refraction in room-scale environments, showcasing its potential as a novel modeling approach capable of solving previously intractable problems beyond the scope of conventional FEM, data-driven techniques, and standard PINNs. As scenario complexity increases, e.g., with additional scatterers and material interfaces, additional kernel functions are required for accurate solutions. Notably, PE-PINN's computation time does not scale with the number of spatial sampling points. For instance, in scenarios 3 vs. 4, 9 vs. 10, and 11 vs. 12 (comparable physical settings and kernel counts), with the default collocation spacing of 0.015 m, the 3D cases involve tens of millions of sampling points, more than two orders of magnitude more than their 2D counterparts, with total training time increasing by less than triple. This property enables PE-PINN to mitigate the curse of dimensionality in numerical modeling \cite{koppen2000curse}. Currently, kernel functions and domain decompositions are manually selected based on prior knowledge of wave physics and environmental properties. Future work will develop automated methods to determine optimal kernel functions and domain decompositions for given scenarios.
\vspace{-2mm}



\section{Related Works}


\subsection{Physics-Informed Neural Networks and Scalability}

The concept of Physics-Informed Neural Networks (PINNs) was seminalized in~\cite{RAISSI2019686}, establishing a deep learning framework for solving forward and inverse problems involving nonlinear partial differential equations (PDEs). By embedding these governing equations directly into the loss function, PINNs enable mesh-free modeling even under regimes of limited supervision. This study integrates a diverse spectrum of physical laws into the training process, covering fundamental dynamics in fluid mechanics, quantum physics, and reaction-diffusion systems.

Despite the flexibility, PINNs are known to suffer from optimization difficulties such as loss term imbalance and slow convergence, especially when multiple constraints are present. Various prior works implemented adaptive weighting and multi-objective balancing schemes to stabilize training~\cite{doi:10.1137/20M1318043,Bischof_2025}. 

Furthermore, to address scalability in complex geometries, domain decomposition strategies have been developed to partition the global solution into manageable subnetworks coupled via interface constraints. Representative approaches, such as XPINNs~\cite{jagtap2020extended} and finite basis PINNs (FBPINNs)~\cite{Moseley_2023}, effectively reduce problem complexity and facilitate multi-scale learning by optimizing smaller, localized subdomains.

Recent neural operator methods, notably the Fourier Neural Operator (FNO)~\cite{li2021fourierneuraloperatorparametric} and DeepONet~\cite{Lu_2021}, learn a mapping from input functions to solution fields, enabling rapid inference across varying PDE instances once trained. While FNO-based approaches excel at amortized inference once trained, they require large volumes of high-fidelity paired data generated by classical solvers. In contrast, PE-PINN is entirely label-free: it optimizes the PDE residual and boundary conditions directly for an instance without any pre-computed training data. 

\subsection{Spectral Bias in High-Frequency Wave Reconstruction}

While PINNs have shown success in various PDEs, applying them to the Helmholtz equation for high-frequency wave propagation remains a significant challenge due to spectral bias, the tendency of neural networks to prioritize learning low-frequency functions~\cite{rahaman2019spectral}. Standard machine learning models struggle to capture the rapid oscillations characteristic of large-scale wavefields. 

To overcome spectral bias, recent research has focused on input encoding and specialized basis functions. Fourier feature mappings have been widely adopted to project input coordinates into higher-dimensional feature spaces, thereby enabling networks to resolve higher frequency components~\cite{tancik2020fourierfeaturesletnetworks, inproceedings}. Subsequent advancements have refined this approach through iterative training strategies that progressively resolve higher frequency bands~\cite{wu2025iterativetrainingphysicsinformedneural}, as well as frameworks combining Fourier features with deep residual stacking architectures~\cite{HOU2026108247}. However, these methods typically rely on generic frequency embeddings, which still force the network to approximate complex wavefronts directly from coordinate inputs.

Beyond global Fourier features, researchers have explored localized basis functions to better represent wave behaviors. A recent study ~\cite{abedi2025gaborenhancedphysicsinformedneuralnetworks} introduced a Gabor-Enhanced PINN that maps coordinates to a custom Gabor coordinate system, effectively absorbing oscillations into the basis functions to simplify learning. While this approach improves upon standard PINNs by localizing frequency representation, it relies on a fixed set of Gabor parameters. Consequently, in scenarios involving complex models, this method still struggles to capture high-frequency variations.

Our work is aligned with the intention of improving PINNs for wave problems, but it diverges in methodology. Instead of relying solely on optimization strategies or generic feature enrichments, we embed physically motivated structures directly into the network architecture. The proposed PE-PINN introduces an envelope transformation layer that explicitly decouples rapidly oscillatory carriers from slowly varying envelopes based on prior physical knowledge of wave propagation, reflection, diffraction, and refraction behaviors. This design effectively bypasses spectral bias by shifting the learning objective to smooth functions, achieving faster convergence and superior accuracy compared to existing baselines.

\section{Conclusion}\label{sec:conclusion}

In this work, we introduced PE-PINN, a new framework that integrates physical knowledge directly into the network architecture, not merely as constraints within the loss function. By incorporating physics-embedded envelope transformation layers and distinct sub-networks to separate incident/scattered fields and spatial domains with varying material properties, PE-PINN shows great promise in resolving the spectral bias and curse of dimensionality that constrained previous PINNs for large-scale wave field reconstruction. The results demonstrated significant efficiency gains with practical implications for wave propagation applications such as wireless systems, acoustics, etc., which are intractable by the existing approaches. 

\textbf{AI Use Disclosure:} We used generative AI tools to assist with the creation, editing, and debugging of software code, as well as with editing the texts to improve readability. All AI-assisted code was reviewed, verified, and tested for correctness by the authors. The authors take full responsibility for the final content of this work, including all text, claims, and artifacts produced with the assistance of generative AI.




\newpage
\appendix
\onecolumn
\section{Wave Field Reconstruction Results using PE-PINN.}
\label{appendix A}

\subsection{Summary on the Evaluation Results}

The evaluations results for PE-PINN in all the studied scenarios are listed in Table~\ref{tab:performance_comparison_rest}. Details on the scenarios settings and more results are presented in the rest of ~\ref{appendix A}.

\begin{table*}[ht!]
    \centering
    \scriptsize             
    \setlength{\tabcolsep}{4pt}
    \caption{Evaluation results on the rest of the scenarios. Note that the benchmarking models cannot converge in reasonable amount of time.}
    \label{tab:performance_comparison_rest}
    \begin{tabular}{ccccccc}
        \toprule
       Scenario No.  & Framework & Time/100 Iters & Time (h:m:s) & Total Iters & $\mathcal{L}_{\text{pde}}$ & $\text{MSE}_{\text{(vs.COMSOL)}}^{\text{After\ norm}}$ \\
        \midrule
        
         1 & PE-PINN & 2.11 s & 00:17:54 & 50 K & 1.11e-01 & 3.38e-03\\
         2 & PE-PINN & 3.03 s & 00:10:20 & 20 K & 5.51e-01 & 5.96e-02 \\
         3 & PE-PINN & 0.99 s & 00:04:15 & 25 K & 7.34e-04 & 9.12e-03 \\
         4 & PE-PINN & 1.40 s & 00:11:54 & 50 K  & 1.84e-03 & N/A \\
         5 & PE-PINN & 2.81 s & 00:23:44 & 50 K & 4.27e-02 & 9.29e-03 \\
         6 & PE-PINN & 2.34 s & 00:19:49 & 50 K & 4.49e-01 & 3.98e-02 \\
         7 & PE-PINN & 5.11 s & 00:44:57 & 50 K & 7.43e-02 & 5.01e-03 \\
         8 & PE-PINN & 10.02 s & 02:47:06 & 100 K & 1.04e-02 & N/A \\
         9 & PE-PINN & 3.58 s & 01:00:09 & 100 K & 1.36e-00 & 7.12e-03 \\
         10 & PE-PINN & 4.91 s & 02:45:10 & 200 K & 1.55e-01 & N/A \\
         11 & PE-PINN & 9.19 s & 03:50:27 & 150 K & 3.42e-00 & 1.06e-02 \\
         12 & PE-PINN & 12.27 s & 10:16:36  & 300 K & 1.53e-01 & N/A \\
        \bottomrule
    \end{tabular}
\end{table*}

\subsection{Detailed Settings and Reconstructed Fields of the Studied Scenarios}

Table~\ref{tab:kernel:appendix} lists all the kernel settings applied to PE-PINN in all the studied scenarios. More information on the configurations for each studied scenario is given below.

\begin{table}[ht!]
    \centering
    \scriptsize             
    \setlength{\tabcolsep}{4pt}
    \caption{Kernel settings in PE-PINN for all studied scenarios.}
    \label{tab:kernel:appendix}
    \begin{tabular}{cccc}
        \toprule
       Scenario No.  & Total kernel number & Plane Wave Kernel $\mathbf{d}_m$ & Spherical Wave Kernel $ \mathbf{x}_m$  \\
        \midrule
         1 & 2 & $(1, 0)$ & $(-2.5, 0)$ \\
         \midrule
         2 & 2 & $(1, 0, 0)$ &  $(-0.5, 0, 0)$ \\
         \midrule
         3 & 1 & N/A & $(-2, 0)$ \\
         \midrule
         4 & 1 & N/A & $(-2, 0, 0)$\\
         \midrule
         5 & 2 & $(\frac{\sqrt{2}}{2}, \frac{\sqrt{2}}{2})$, $(-\frac{\sqrt{2}}{2}, \frac{\sqrt{2}}{2})$  & N/A \\
         \midrule
         6 & 2 & N/A & $(-2, 0)$, $(1, 0)$ \\
         \midrule
         7 & 4 & N/A & $(-2, 0)$, $(1, 0)$, $(-0.5, 0.5)$, $(-0.5, -0.5)$ \\
         \midrule
         \multirow{2}{*}{8} & \multirow{2}{*}{6} & \multirow{2}{*}{N/A} & $(-2, 0, 0), (-0.5, -0.5, 0.5), (-0.5, 0.5, 0.5),$ \\
                                                    & & & $(1, 0, 0), (-0.5, 0.5, -0.5), (-0.5, -0.5, -0.5)$ \\
        \midrule
         9 & 3 & N/A & $(-2, 0)$, $(1, 0)$,$(-7, 0)$ \\
         \midrule
         10 & 3 & N/A  & $(-2, 0, 0)$, $(1, 0, 0)$, $(-7, 0, 0)$ \\
         \midrule
         \multirow{2}{*}{11} & \multirow{2}{*}{8} & \multirow{2}{*}{$(1, 0)$, $(-1, 0)$} & $(-2, 0)$, $(1.1, 0)$,$(1.5, 0)$,\\
                                                                                    & & & $(-7, 0)$, $(7.9, 0)$, $(-1.4, 0)$ \\
         \midrule
         \multirow{2}{*}{12} & \multirow{2}{*}{8} & \multirow{2}{*}{$(1, 0, 0)$, $(-1, 0, 0)$} & $(-2, 0, 0)$,$(1.1, 0,0)$,$(1.5, 0,0)$, \\
                                                                                            & & & $(-7, 0, 0)$, $(7.9, 0, 0)$, $(-1.4, 0, 0)$\\
        \bottomrule
    \end{tabular}
\end{table}

\begin{figure}[ht!]
    \centering
    \includegraphics[width=0.75\linewidth]{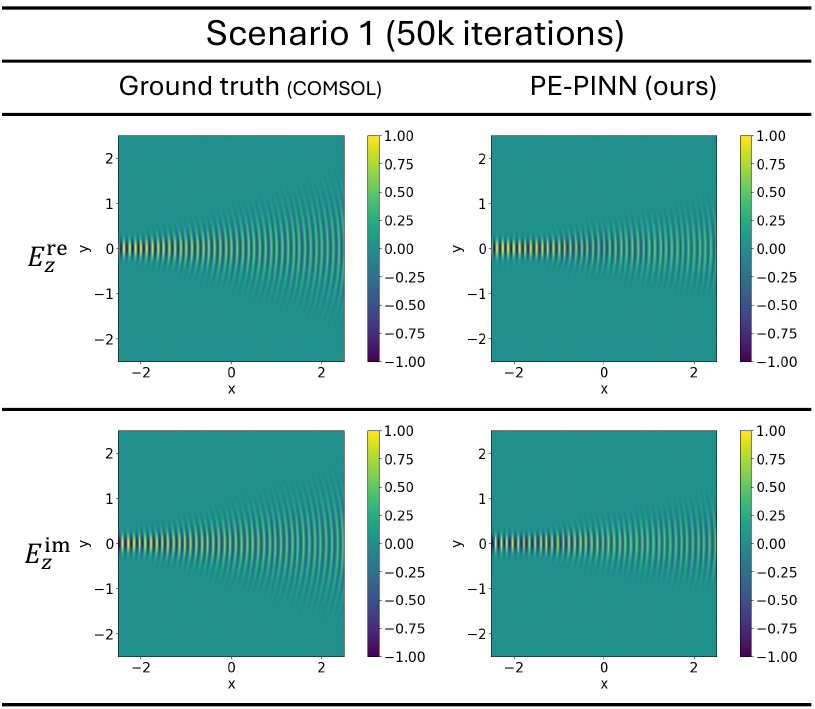}
    \caption{Wave field reconstruction results for 2D plane wave source (Scenario 1).}
  \label{fig:scenario1:appendix}
\end{figure}

\noindent\textbf{Scenario 1:}
The computational domain for the 2D free-space case (Scenario 1) is defined as $\Omega = [-2.5~\text{m}, 2.5~\text{m}]^2$. An incident source is introduced at the boundary $(-2.5, 0)$ with a Gaussian profile $E_z^\text{Re}(\mathbf{x}) = \exp(- \frac{y^2}{0.2^2})$. While the wavefront is initially planar at the source boundary, the finite transverse width introduces diffraction effects, causing the beam to diverge and the wavefronts to gradually evolve from planar to cylindrical in the far field. To accurately reconstruct this complex propagation behavior, the PE-PINN integrates two kernel functions. The plane wave kernel $\mathbf{d}_1=(1,0)$ and the spherical wave kernel $\mathbf{x}_1=(-2.5,0)$ .
\begin{equation}
E_{z}(\mathbf{x}) \approx \mathcal{N}_{\text{tot}}(\mathbf{x}; \Psi_{\mathbf{d}_1}, \Psi_{\mathbf{x}_1}).
\end{equation}
Fig.~\ref{fig:scenario1:appendix} shows the reconstruction of the wave field in Scenario 1.

\begin{figure}[ht!]
    \centering
    \includegraphics[width=\linewidth]{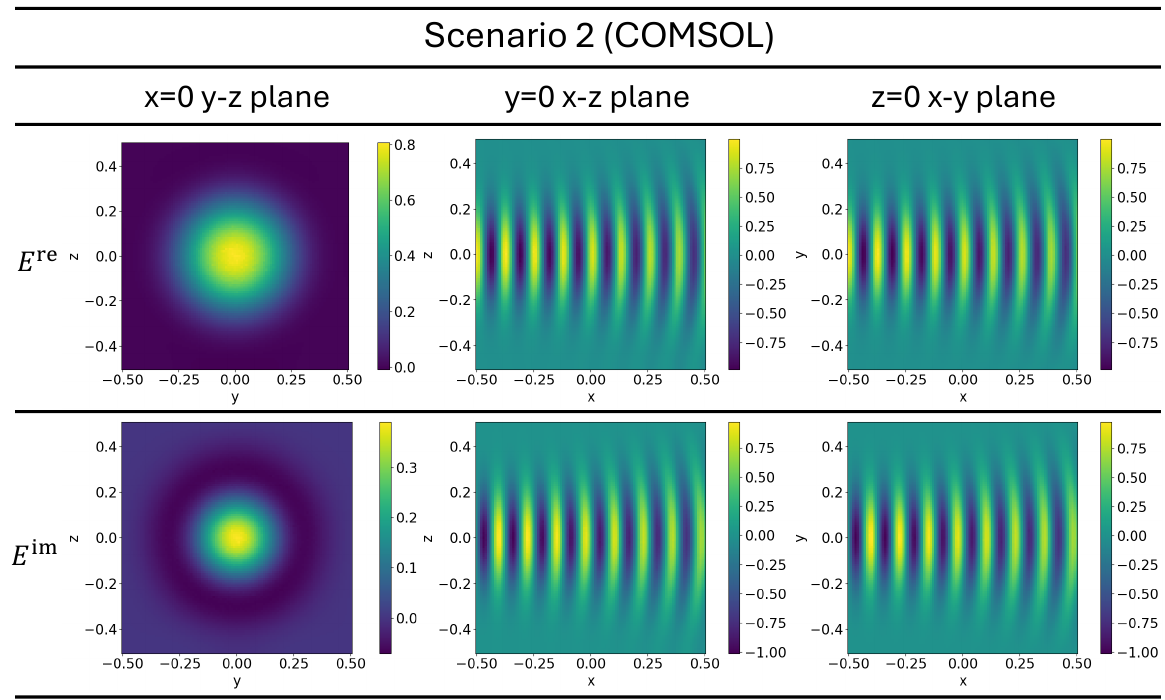}
    \caption{Ground truth for 3D plane wave source free-space propagation (Scenario 2).}
  \label{fig:scenario2:ground_truth:appendix}
\end{figure}

\begin{figure}[ht!]
    \centering
    \includegraphics[width=\linewidth]{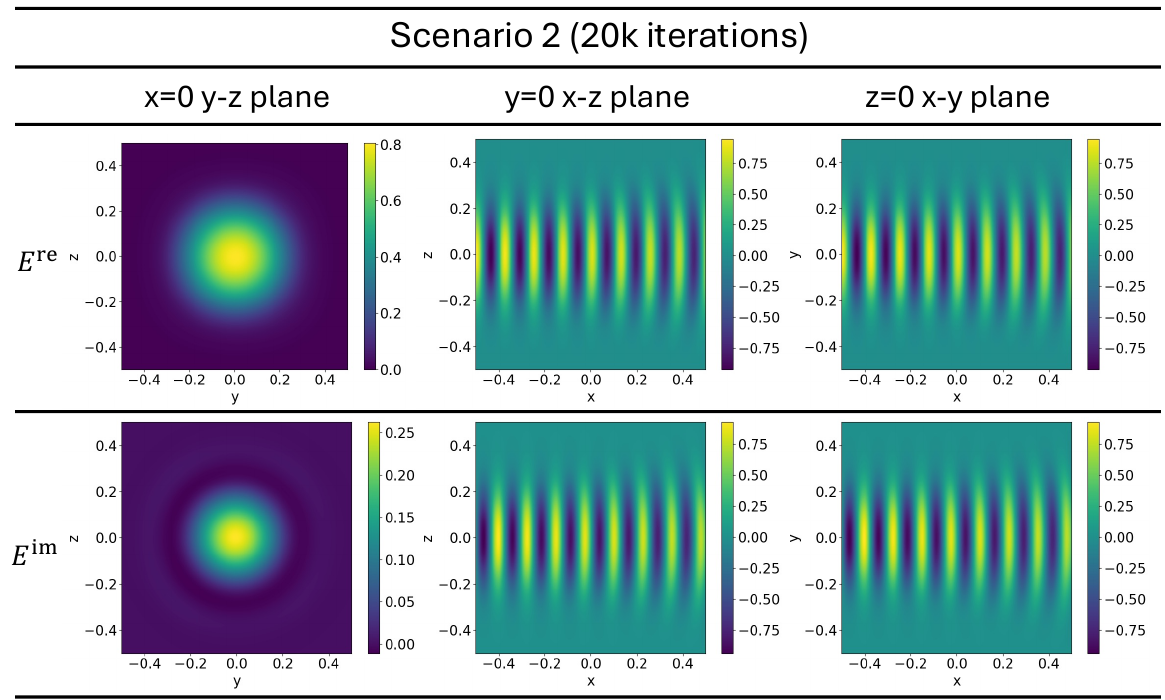}
    \caption{Wave field reconstruction for 3D plane wave source free-space propagation (Scenario 2).}
  \label{fig:scenario2:appendix}
\end{figure}

\noindent\textbf{Scenario 2:}
To facilitate validation against COMSOL benchmarks, the computational domain for the 3D free-space case is scaled down to $\Omega = [-0.5~\text{m}, 0.5~\text{m}]^3$. An incident source is introduced at the boundary $(-0.5, 0, 0)$ with a Gaussian profile $E(\mathbf{x}) = \exp\left(- \frac{y^2+z^2}{0.2^2}\right)$.  To accurately reconstruct complex propagation behavior, the PE-PINN integrates two kernel functions: the plane wave kernel $\mathbf{d}_1=(1,0,0)$ and the spherical wave kernel $\mathbf{x}_1=(-0.5, 0, 0)$.
\begin{equation}
E(\mathbf{x}) \approx \mathcal{N}_{\text{tot}}(\mathbf{x}; \Psi_{\mathbf{d}_1}, \Psi_{\mathbf{x}_1}).
\end{equation}
Fig.~\ref{fig:scenario2:ground_truth:appendix} and Fig.~\ref{fig:scenario2:appendix} show the FEM reference computed from COMSOL and the results from PE-PINN in Scenario 2, respectively. Note that COMSOL already requires more than 100 GB of RAM in this small-scale 3D scenario. Regarding computational time, the COMSOL simulation involves 2 minutes for meshing and 10 minutes for the solution phase, amounting to a total of 12 minutes. In contrast, the PE-PINN approach completes the task in 10 minutes and 20 seconds (Table.~\ref{tab:performance_comparison_rest}) for this case. 


\begin{figure}[ht!]
    \centering
    \includegraphics[width=0.75\linewidth]{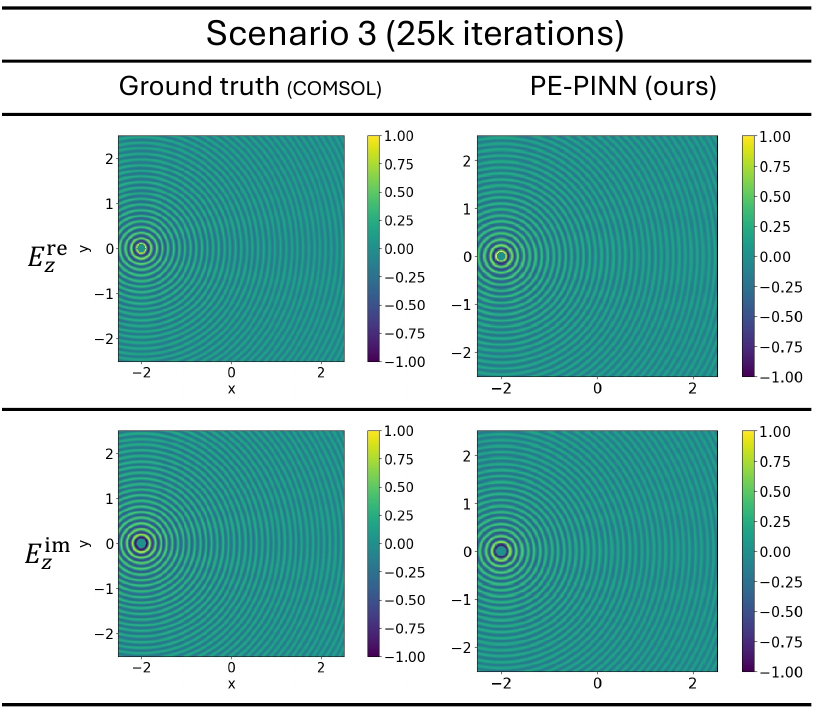}
    \caption{Wave field reconstruction comparison for 2D spherical source free space propagation (scenario 3).}
  \label{fig:scenario3:appendix}
\end{figure}

\noindent\textbf{Scenario 3:}
The computational domain for this free-space propagation scenario is defined as $\Omega = [-2.5~\text{m}, 2.5~\text{m}]^2$. In this case, no scatterers or reflectors are present within the domain. A single network $\mathcal{N}_{\text{tot}}$ is utilized. This network is dedicated to modeling using the spherical wave kernel $\mathbf{x}_1=(-2, 0)$. As the wave propagates without obstruction, the total field is governed solely by the primary radiation, requiring no additional scattering kernels.
\begin{equation}
E_{z}(\mathbf{x}) \approx \mathcal{N}_{\text{tot}}(\mathbf{x}; \Psi_{\mathbf{x}_1}).
\end{equation}
Fig.~\ref{fig:scenario3:appendix} shows the reconstruction of the wave field in Scenario 3. 

\begin{figure}[ht!]
    \centering
    \includegraphics[width=\linewidth]{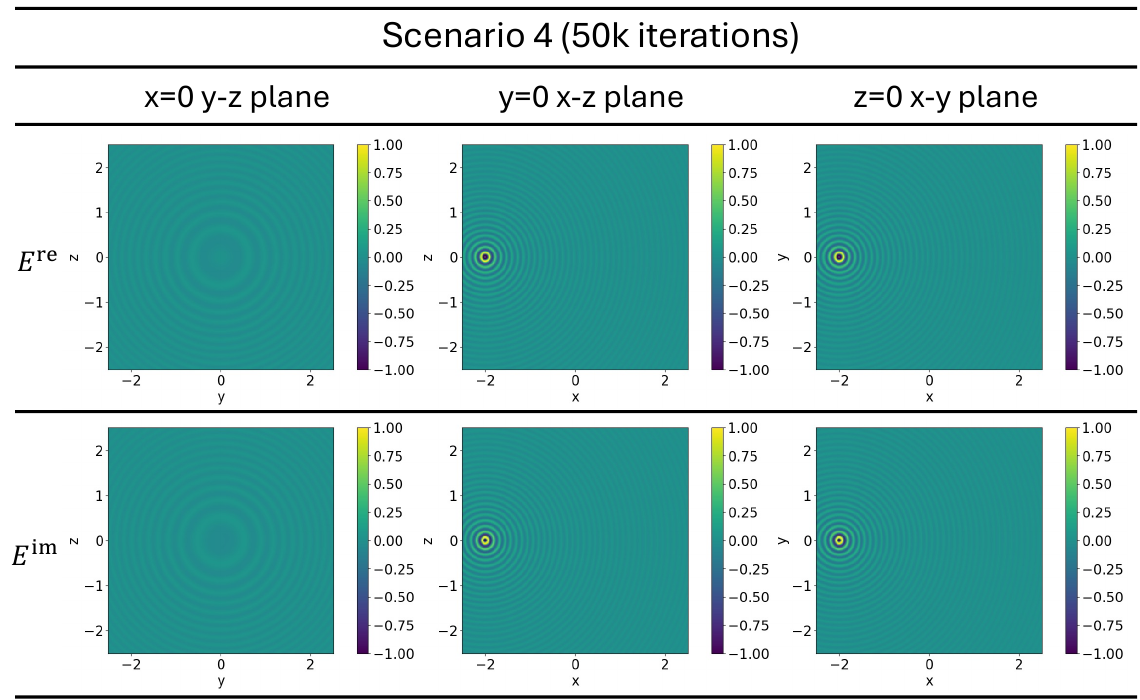}
    \caption{Wave field reconstruction for 3D point source propogation (scenario 4).}
  \label{fig:scenario4:appendix}
\end{figure}

\noindent\textbf{Scenario 4:} 
The computational domain for this 3D free-space propagation scenario is extended to $\Omega = [-2.5~\text{m}, 2.5~\text{m}]^3$. Consistent with the 2D case, no scatterers or reflectors are present within the domain. To accurately reconstruct the wave field, a single network $\mathcal{N}_{\text{tot}}$ is utilized. This network is dedicated to modeling the field using the spherical wave kernel $\mathbf{x}_1=(-2, 0, 0)$. As the wave propagates without obstruction, the total field is governed solely by the primary radiation, requiring no additional scattering kernels.
\begin{equation}
E(\mathbf{x}) \approx \mathcal{N}_{\text{tot}}(\mathbf{x}; \Psi_{\mathbf{x}_1}).
\end{equation}
Fig.~\ref{fig:scenario4:appendix} shows the reconstructed wave field in Scenario 4. 

\begin{figure}[ht!]
    \centering
    \includegraphics[width=0.75\linewidth]{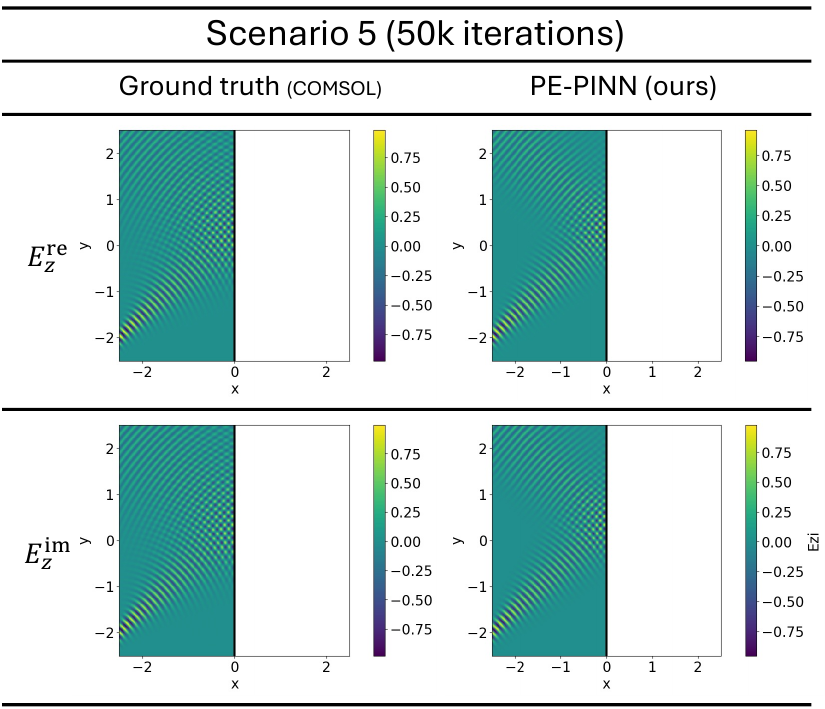}
    \caption{Wave field reconstruction comparison for 2D diagonal plane wave source reflection (Scenario 5).}
  \label{fig:scnario5:appendix}
\end{figure}

\noindent\textbf{Scenario 5:} The computational domain for this reflection scenario is defined as $\Omega = [-2.5~\text{m}, 2.5~\text{m}]^2$. A vertical PEC reflector is located at $x=0$. The incident field is generated by a diagonal Gaussian source ($E_z^\text{Re}(\textbf{x}) = \exp(- \frac{(y + 2.0)^2}{0.2^2}) \cos( k y \cos(\frac{\pi}{4}))$, $E_z^\text{Im}(\textbf{x}) = -\exp(- \frac{(y + 2.0)^2}{0.2^2}) \sin\left( k y \cos(\frac{\pi}{4}\right))$). Since the reflected field propagates away from the source boundary and does not interfere with the incident condition, a single unified network $\mathcal{N}_{\text{tot}}$ is utilized to reconstruct the entire wave field. This network integrates two plane wave kernels to capture the interference pattern: the incident kernel $\mathbf{d}_1 = (\frac{\sqrt{2}}{2}, \frac{\sqrt{2}}{2})$, and the reflection kernel $\mathbf{d}_2 = (-\frac{\sqrt{2}}{2}, \frac{\sqrt{2}}{2})$.
\begin{equation}
E_{z}(\mathbf{x}) \approx \mathcal{N}_{\text{tot}}(\mathbf{x}; \Psi_{\mathbf{d}_1}, \Psi_{\mathbf{d}_2}).
\end{equation}
Fig.~\ref{fig:scnario5:appendix} shows the reconstructed wave field in Scenario 5. 

\begin{figure}[ht!]
    \centering
    \includegraphics[width=0.75\linewidth]{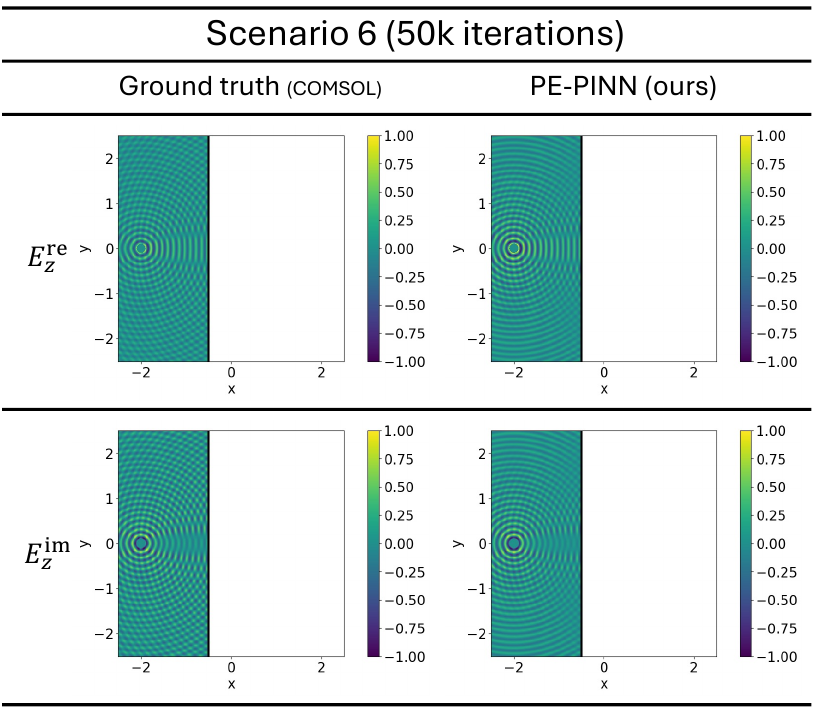}
    \caption{Wave field reconstruction comparison for 2D spherical source reflection (scenario 6).}
  \label{fig:scenario6:appendix}
\end{figure}

\noindent\textbf{Scenario 6:}
The computational domain for this reflection scenario is defined as $\Omega = [-2.5~\text{m}, 2.5~\text{m}]^2$. A vertical PEC reflector is located at $x=-0.5~\text{m}$. The dual-network architecture consisting of a $\mathcal{N}_{\text{inc}}$ and $\mathcal{N}_{\text{sct}}$ is used. The source network $\mathcal{N}_{\text{inc}}$ is dedicated to modeling the incident field using the spherical wave kernel $\mathbf{x}_1=(-2, 0)$. $\mathcal{N}_{\text{sct}}$ is designed to capture the pure reflection phenomena by exclusively utilizing the image source kernel $\mathbf{x}_2=(1, 0)$. This kernel corresponds to the mirror image of the primary source with respect to the PEC boundary. 
\begin{equation}
E_{z}(\mathbf{x}) \approx \mathcal{N}_{\text{inc}}(\mathbf{x}; \Psi_{\mathbf{x}_1}) + \mathcal{N}_{\text{sct}}(\mathbf{x}; \Psi_{\mathbf{x}_2}).
\end{equation}
Fig.~\ref{fig:scenario6:appendix} shows the reconstruction of the wave field in Scenario 6. 

\begin{figure}[ht!]
    \centering
    \includegraphics[width=\linewidth]{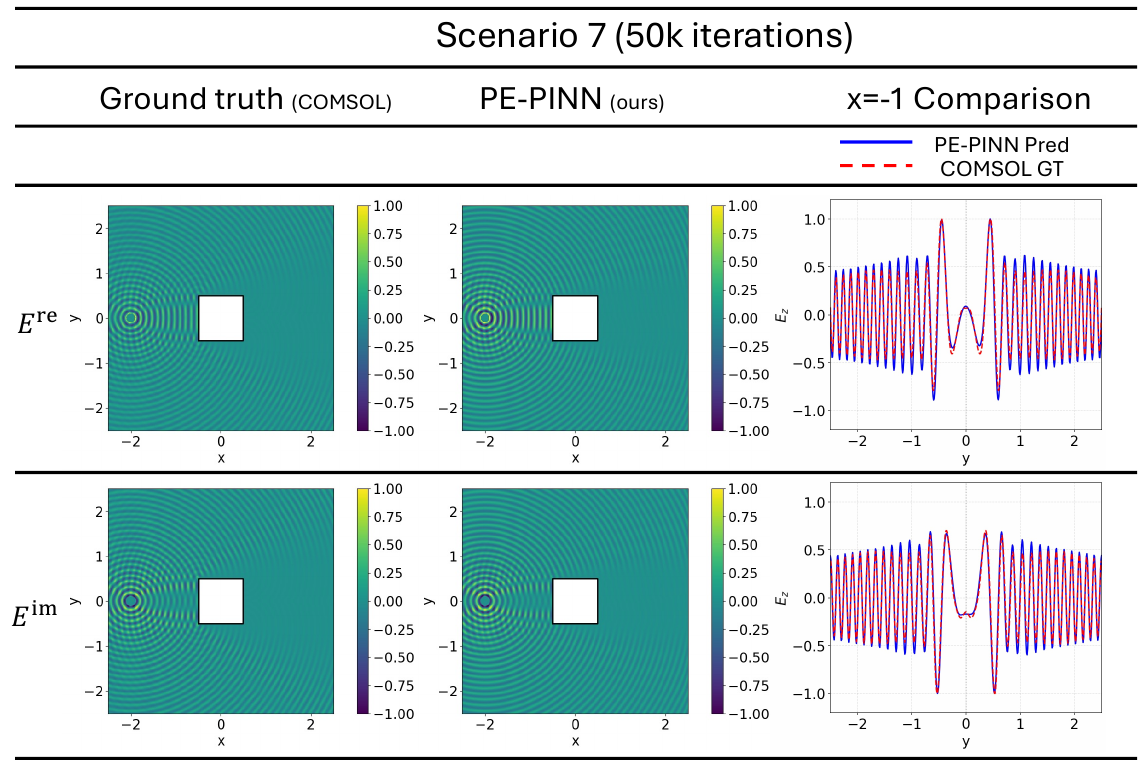}
    \caption{Wave field reconstruction results for 2D spherical source with diffraction (Scenario 7).}
  \label{fig:psdiff-ez:scenario 7:appendix}
\end{figure}

\noindent\textbf{Scenario 7:}  The computational domain for the 2D case (Scenario 7) is defined as $\Omega = [-2.5~\text{m}, 2.5~\text{m}]^2$. A square PEC scatterer with a side length of $1~\text{m}$ is centered at the origin $(0,0)$. To accurately reconstruct the wave field, the dual-network architecture consisting of a $\mathcal{N}_{\text{inc}}$ and $\mathcal{N}_{\text{sct}}$ is used in this scenario. The source network $\mathcal{N}_{\text{inc}}$ is dedicated to modeling the incident field using the spherical wave kernel $\mathbf{x}_1=(-2, 0)$. $\mathcal{N}_{\text{sct}}$ captures the complex scattering phenomena by integrating the primary source kernel $\mathbf{x}_1=(-2, 0)$ alongside the reflection kernel $\mathbf{x}_2=(1, 0)$ (corresponding to the image source) and diffraction kernels $\mathbf{x}_3=(-0.5, 0.5)$, $\mathbf{x}_4=(-0.5, -0.5)$ (originating from the front corners). Crucially, the primary kernel $\Psi_{\mathbf{x}_1}$ is deliberately retained in the basis of $\mathcal{N}_{\text{sct}}$ to address the shadowing effect behind the PEC square.
\begin{equation}
    E_{z}(\mathbf{x}) \approx \mathcal{N}_{\text{inc}}(\mathbf{x}; \Psi_{\mathbf{x}_1}) + \mathcal{N}_{\text{sct}}(\mathbf{x}; \Psi_{\mathbf{x}_1}, \Psi_{\mathbf{x}_2}, \Psi_{\mathbf{x}_3}, \Psi_{\mathbf{x}_4}).
\end{equation}
Fig.~\ref{fig:psdiff-ez:scenario 7:appendix} shows the reconstruction of the wave field in Scenario 7.

\begin{figure}[ht!]
    \centering
    \includegraphics[width=\linewidth]{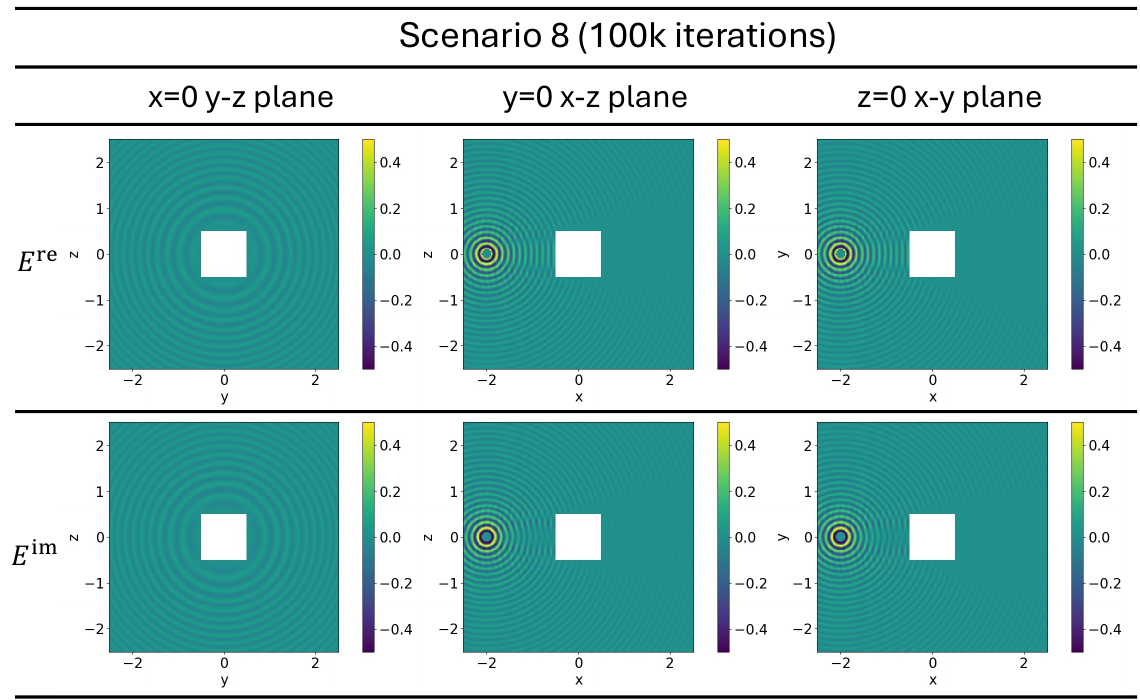}
    \caption{Field reconstruction results ($yz$, $xz$, and $xy$ cross sections) for 3D spherical source with diffraction (Scenario 8).}
  \label{fig:psdiff3d:scenario 8:appendix}
\end{figure}

\noindent\textbf{Scenario 8:} As a direct extension of the 2D scenario, the computational domain for the 3D case (Scenario 8) is expanded to $\Omega = [-2.5~\text{m}, 2.5~\text{m}]^3$. A cubic PEC scatterer with a side length of $1~\text{m}$ is centered at the origin $(0,0,0)$. To accurately reconstruct the volumetric wave field, we employ the same dual-network architecture consisting of $\mathcal{N}_{\text{inc}}$ and $\mathcal{N}_{\text{sct}}$. The source network $\mathcal{N}_{\text{inc}}$ is dedicated to modeling the incident field using the spherical wave kernel $\mathbf{x}_1=(-2, 0, 0)$. $\mathcal{N}_{\text{sct}}$ captures the complex scattering phenomena by integrating the primary source kernel $\Psi_{\mathbf{x}_1}$ alongside the reflection kernel $\mathbf{x}_2=(1, 0, 0)$ (corresponding to the image source). The diffraction kernels are extended from the two 2D corners to the four endpoints of the cube's front vertical edges: $\mathbf{x}_3=(-0.5, 0.5, 0.5)$, $\mathbf{x}_4=(-0.5, -0.5, 0.5)$, $\mathbf{x}_5=(-0.5, 0.5, -0.5)$, and $\mathbf{x}_6=(-0.5, -0.5, -0.5)$. Consistent with the 2D strategy, the primary kernel $\Psi_{\mathbf{x}_1}$ is deliberately retained in the basis of $\mathcal{N}_{\text{sct}}$ to explicitly address the shadowing effect behind the PEC cube.
\begin{equation}
    E(\mathbf{x}) \approx \mathcal{N}_{\text{inc}}(\mathbf{x}; \Psi_{\mathbf{x}_1}) + \mathcal{N}_{\text{sct}}(\mathbf{x}; \Psi_{\mathbf{x}_1}, \Psi_{\mathbf{x}_2}, \Psi_{\mathbf{x}_3}, \Psi_{\mathbf{x}_4}, \Psi_{\mathbf{x}_5}, \Psi_{\mathbf{x}_6}).
\end{equation}
Fig.~\ref{fig:psdiff3d:scenario 8:appendix} shows the reconstruction of the wave field in Scenario 8. 

\begin{figure}[ht!]
    \centering
    \includegraphics[width=\linewidth]{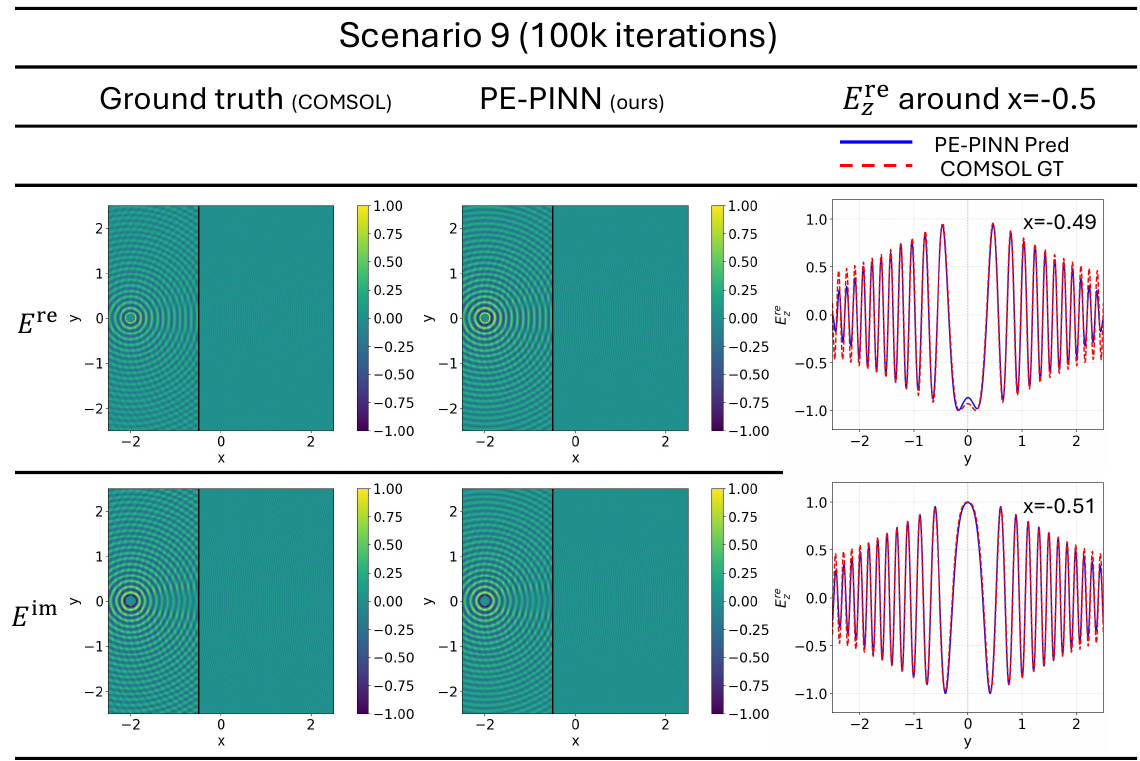}
    \caption{Wave field reconstruction results for 2D spherical source with refraction (Scenario 9). }
  \label{fig:psdiff2d:scenario 9:appendix}
\end{figure}

\noindent\textbf{Scenario 9:} 
The computational domain for this refraction scenario is defined as $\Omega = [-2.5~\text{m}, 2.5~\text{m}]^2$. The domain is divided by a vertical interface at $x=-0.5~\text{m}$ into two distinct regions: an air region ($x < -0.5$) with $\epsilon_r=1$ and a dielectric medium ($x > -0.5$) with $\epsilon_r=9$. To accurately reconstruct the wave field across this heterogeneous medium, a multi-network architecture comprising $\mathcal{N}_{\text{inc}}$, $\mathcal{N}_{\text{sct,L}}$, and $\mathcal{N}_{\text{sct,R}}$ is utilized. In the air region ($x < -0.5$), $\mathcal{N}_{\text{inc}}$ is dedicated to modeling the incident field using the source kernel $\mathbf{x}_1=(-2, 0)$, while $\mathcal{N}_{\text{sct,L}}$ captures the reflected wave using the image source kernel $\mathbf{x}_2=(1, 0)$. In the dielectric region ($x > -0.5$), $\mathcal{N}_{\text{sct,R}}$ models the transmitted wave propagation using a virtual source kernel $\mathbf{x}_3=(-7, 0)$, which lies on the refraction caustic and accounts for the refractive index change. It is worth noting that since the incident field is excluded from this region, $\mathcal{N}_{\text{sct,R}}$ solely constitutes the total field. However, for notational consistency, we retain the 'sct' label. This convention, designating the computed total field as 'sct' specifically in regions where the incident field is absent, is applied throughout the subsequent analysis. 
\begin{equation}
E_{z}(\mathbf{x}) \approx 
\begin{cases} 
\mathcal{N}_{\text{inc}}(\mathbf{x}; \Psi_{\mathbf{x}_1}) + \mathcal{N}_{\text{sct,L}}(\mathbf{x}; \Psi_{\mathbf{x}_2}), & x < -0.5 \\
\mathcal{N}_{\text{sct,R}}(\mathbf{x}; \Psi_{\mathbf{x}_3}), & x > -0.5
\end{cases}
\end{equation}
Fig.~\ref{fig:psdiff2d:scenario 9:appendix} shows the reconstruction of the wave field in Scenario 9. To verify the accuracy across the material discontinuity, detailed real part comparisons are provided adjacent to the refraction interface at $x=-0.51$ and $x=-0.49$.

\begin{figure}[ht!]
    \centering
    \includegraphics[width=\linewidth]{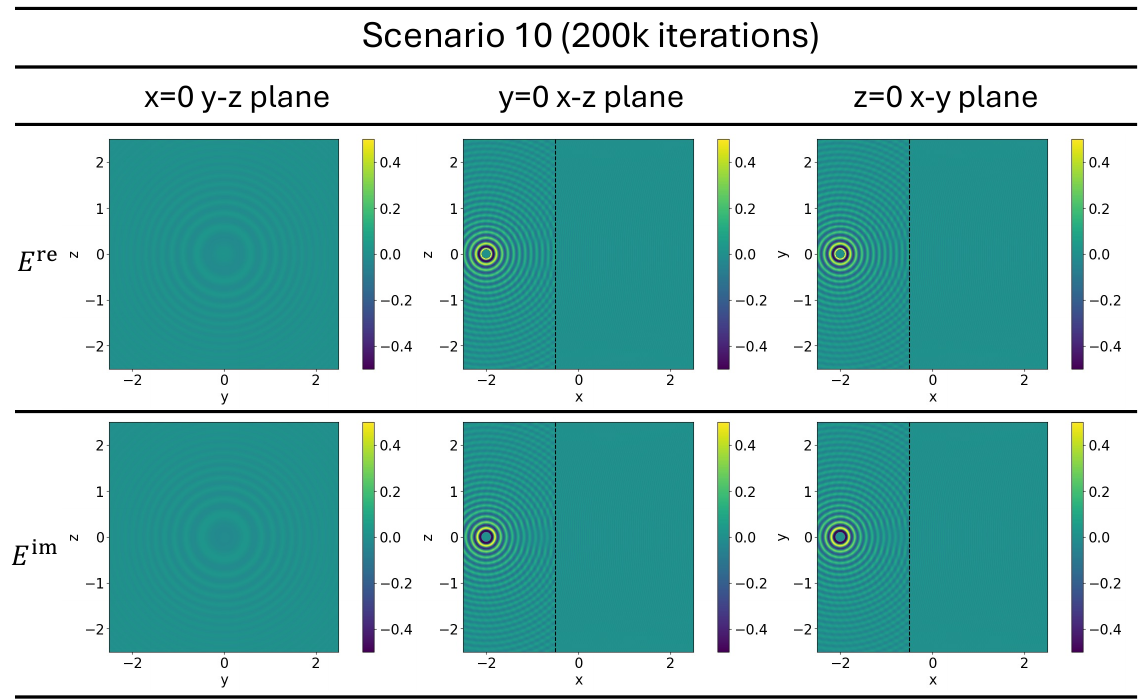}
    \caption{Wave field reconstruction results ($yz$, $xz$, and $xy$ cross sections) for 3D spherical source with refraction (Scenario 10).}
  \label{fig:psdiff3d:scenario 10:appendix}
\end{figure}

\noindent\textbf{Scenario 10:}
As a direct extension of the 2D refraction scenario, the computational domain for the 3D case is expanded to $\Omega = [-2.5~\text{m}, 2.5~\text{m}]^3$. The domain is divided by a vertical planar interface at $x=-0.5~\text{m}$ into two distinct regions: an air region ($x < -0.5$) with $\epsilon_r=1$ and a dielectric medium ($x > -0.5$) with $\epsilon_r=9$. To accurately reconstruct the wave field across this heterogeneous medium, we utilize the same multi-network architecture comprising $\mathcal{N}_{\text{inc}}$, $\mathcal{N}_{\text{sct,L}}$, and $\mathcal{N}_{\text{sct,R}}$. In the air region ($x < -0.5$), $\mathcal{N}_{\text{inc}}$ is dedicated to modeling the incident field using the spherical wave kernel $\mathbf{x}_1=(-2, 0, 0)$, while $\mathcal{N}_{\text{sct,L}}$ captures the reflected wave using the image source kernel $\mathbf{x}_2=(1, 0, 0)$. In the dielectric region ($x > -0.5$), $\mathcal{N}_{\text{sct,R}}$ models the transmitted wave propagation using a virtual source kernel $\mathbf{x}_3=(-7, 0, 0)$, which effectively accounts for the refractive index change in 3D space.
\begin{equation}
    E(\mathbf{x}) \approx 
    \begin{cases} 
    \mathcal{N}_{\text{inc}}(\mathbf{x}; \Psi_{\mathbf{x}_1}) + \mathcal{N}_{\text{sct,L}}(\mathbf{x}; \Psi_{\mathbf{x}_2}), & x < -0.5 \\
    \mathcal{N}_{\text{sct,R}}(\mathbf{x}; \Psi_{\mathbf{x}_3}), & x > -0.5
    \end{cases}
\end{equation}
Fig.~\ref{fig:psdiff3d:scenario 10:appendix} shows the reconstruction of the wave field in Scenario 10.

\begin{figure}[ht!]
    \centering
    \includegraphics[width=0.75\linewidth]{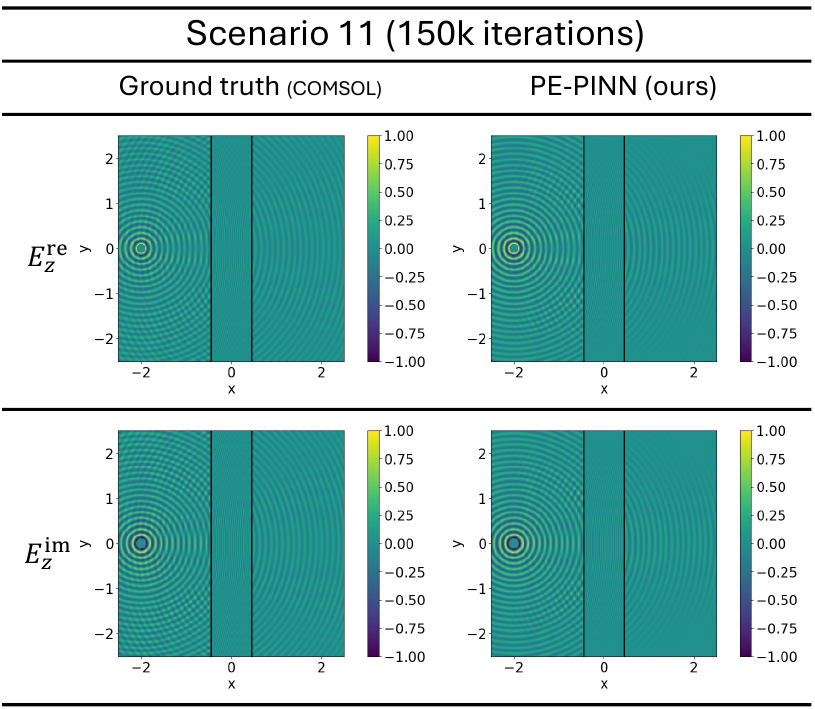}
    \caption{Wave field reconstruction results for 2D point source strip refraction (Scenario 11).}
  \label{fig:scenario11:appendix}
\end{figure}

\noindent\textbf{Scenario 11:}
The computational domain for this dielectric strip scenario is defined as $\Omega = [-2.5~\text{m}, 2.5~\text{m}]^2$. The domain is characterized by a central dielectric strip ($\epsilon_r=9$) located within $-0.45~\text{m} \le x \le 0.45~\text{m}$, surrounded by air regions ($\epsilon_r=1$) on both sides. To accurately reconstruct the wave field across this multi-layered medium, a composite architecture comprising four networks is utilized: $\mathcal{N}_{\text{inc}}$, $\mathcal{N}_{\text{sct,L}}$, $\mathcal{N}_{\text{sct,M}}$, and $\mathcal{N}_{\text{sct,R}}$.
In the left air region ($x < -0.45$), $\mathcal{N}_{\text{inc}}$ models the incident field with the source kernel $\mathbf{x}_1=(-2, 0)$, while $\mathcal{N}_{\text{sct,L}}$ captures the reflection from the strip using two spherical wave kernels at $\mathbf{x}_2=(1.1, 0)$ and $\mathbf{x}_3=(1.5, 0)$. In the central dielectric region ($-0.45 \le x \le 0.45$), $\mathcal{N}_{\text{sct,M}}$ is designed to capture the complex internal standing waves resulting from multiple reflections; it integrates two spherical kernels at $\mathbf{x}_4=(-7, 0)$ and $\mathbf{x}_5=(7.9, 0)$ alongside two plane wave kernels with directions $\mathbf{d}_1=(1, 0)$ and $\mathbf{d}_2=(-1, 0)$. Finally, in the right air region ($x > 0.45$), $\mathcal{N}_{\text{sct,R}}$ models the transmitted wave using a single spherical kernel at $\mathbf{x}_6=(-1.4, 0)$.

\begin{equation}
E_{z}(\mathbf{x}) \approx 
\begin{cases} 
\mathcal{N}_{\text{inc}}(\mathbf{x}; \Psi_{\mathbf{x}_1}) + \mathcal{N}_{\text{sct,L}}(\mathbf{x}; \Psi_{\mathbf{x}_2}, \Psi_{\mathbf{x}_3}), & x < -0.45 \\
\mathcal{N}_{\text{sct,M}}(\mathbf{x}; \Psi_{\mathbf{x}_4}, \Psi_{\mathbf{x}_5}, \Psi_{\mathbf{d}_1}, \Psi_{\mathbf{d}_2}), & -0.45 \le x \le 0.45 \\
\mathcal{N}_{\text{sct,R}}(\mathbf{x}; \Psi_{\mathbf{x}_6}), & x > 0.45
\end{cases}
\end{equation}
Fig.~\ref{fig:scenario11:appendix} shows the reconstruction of the wave field in Scenario 11. 

\noindent\textbf{Scenario 12:} As a direct extension of the 2D dielectric strip scenario, the computational domain for the 3D case is expanded to $\Omega = [-2.5~\text{m}, 2.5~\text{m}]^3$. The domain features a central dielectric strip ($\epsilon_r=9$) spanning the volumetric region $-0.45~\text{m} \le x \le 0.45~\text{m}$, sandwiched between air regions ($\epsilon_r=1$). To accurately reconstruct the volumetric wave field, we utilize the same composite architecture comprising four networks: $\mathcal{N}_{\text{inc}}$, $\mathcal{N}_{\text{sct,L}}$, $\mathcal{N}_{\text{sct,M}}$, and $\mathcal{N}_{\text{sct,R}}$.

\begin{figure}[ht!]
    \centering
    \includegraphics[width=\linewidth]{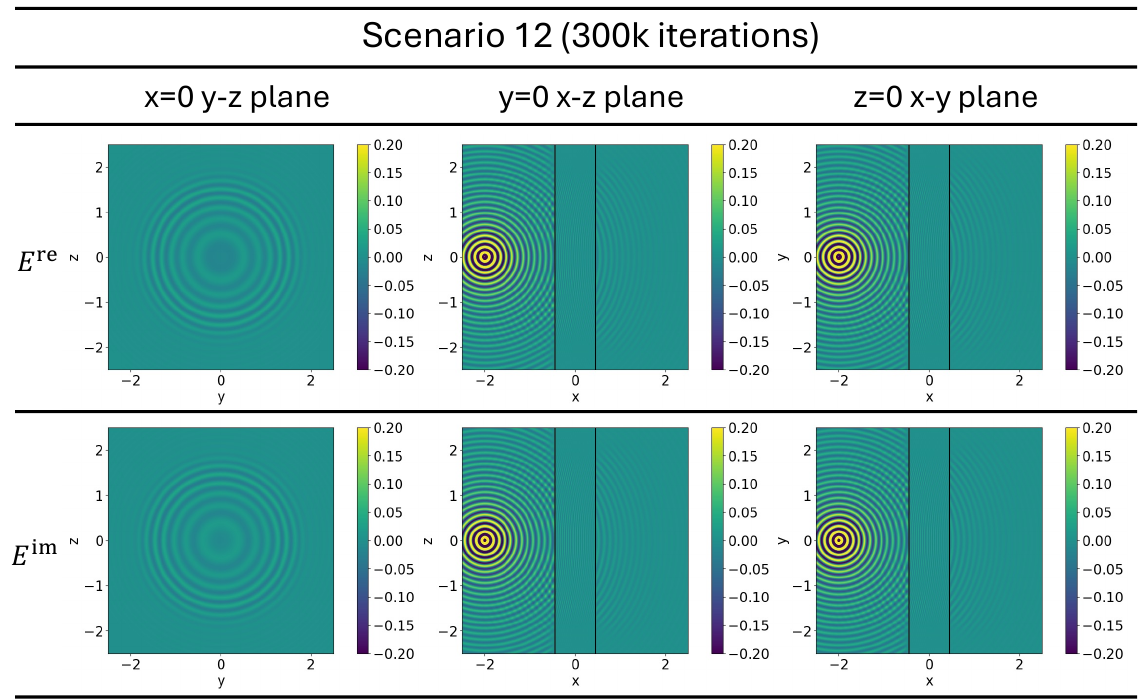}
    \caption{Wave field reconstruction results ($yz$, $xz$, and $xy$ cross sections) for 3D point source strip refraction (Scenario 12).}
  \label{fig:scenario12:appendix}
\end{figure}

In the left air region ($x < -0.45$), $\mathcal{N}_{\text{inc}}$ models the incident field with the source kernel $\Psi_{\mathbf{x}_1}$ centered at $\mathbf{x}_1=(-2, 0, 0)$, while $\mathcal{N}_{\text{sct,L}}$ captures the reflection from the strip using two spherical wave kernels at $\mathbf{x}_2=(1.1, 0, 0)$ and $\mathbf{x}_3=(1.5, 0, 0)$. In the central dielectric region ($-0.45 \le x \le 0.45$), $\mathcal{N}_{\text{sct,M}}$ is designed to capture the complex internal standing waves; it integrates two spherical kernels at $\mathbf{x}_4=(-7, 0, 0)$ and $\mathbf{x}_5=(7.9, 0, 0)$ alongside two plane wave kernels with propagation directions $\mathbf{d}_1=(1, 0, 0)$ and $\mathbf{d}_2=(-1, 0, 0)$. Finally, in the right air region ($x > 0.45$), $\mathcal{N}_{\text{sct,R}}$ models the transmitted wave using a single spherical kernel at $\mathbf{x}_6=(-1.4, 0, 0)$.

\begin{equation}
    E(\mathbf{x}) \approx 
    \begin{cases} 
    \mathcal{N}_{\text{inc}}(\mathbf{x}; \Psi_{\mathbf{x}_1}) + \mathcal{N}_{\text{sct,L}}(\mathbf{x}; \Psi_{\mathbf{x}_2}, \Psi_{\mathbf{x}_3}), & x < -0.45 \\
    \mathcal{N}_{\text{sct,M}}(\mathbf{x}; \Psi_{\mathbf{x}_4}, \Psi_{\mathbf{x}_5}, \Psi_{\mathbf{d}_1}, \Psi_{\mathbf{d}_2}), & -0.45 \le x \le 0.45 \\
    \mathcal{N}_{\text{sct,R}}(\mathbf{x}; \Psi_{\mathbf{x}_6}), & x > 0.45
    \end{cases}
\end{equation}
Fig.~\ref{fig:scenario12:appendix} shows the reconstruction of the wave field in Scenario 12. 

\section{Loss Weighting Strategy}
\label{appendix B}

The composite loss function in Eq. \eqref{eq:total_loss} consists of multiple terms with physically distinct units and magnitudes. To mitigate the optimization stiffness caused by gradient imbalance, we employ a flexible weighting strategy that adapts to the problem complexity.

\noindent \textbf{Fixed Weighting:} For scenarios with stable convergence characteristics (e.g., homogeneous media), we utilize fixed weighting coefficients. These coefficients $\lambda_i$ are pre-determined empirically or calibrated based on the initial magnitude of the residuals to ensure that all loss components effectively contribute to the gradient updates.

\noindent \textbf{Adaptive Weighting:} For complex scenarios, we utilize a simplified Gradient Normalization strategy \cite{doi:10.1137/20M1318043}. Our approach computes the global $L_2$ norm of the gradients with respect to all network parameters $\Theta$. This simplification reduces the computational overhead while capturing the overall contribution of each loss term. The adaptive weights are updated using the Exponential Moving Average (EMA) to stabilize the training dynamics:
\begin{equation}
\begin{aligned}
    G_i^{(t)} &= \|\nabla_{\Theta} \mathcal{L}_i^{(t)}\|_2, \\
    \bar{G}_i^{(t)} &= \alpha \bar{G}_i^{(t-1)} + (1 - \alpha) G_i^{(t)}, \\
    \lambda_i^{(t)} &= \frac{\max_j \{ G_j^{(t)} \}}{\bar{G}_i^{(t)} + \epsilon},
\end{aligned}
\end{equation}
where $G_i^{(t)}$ is the instantaneous global gradient norm and $\bar{G}_i^{(t)}$ is its smoothed history. This formulation ensures that all physical constraints contribute equivalently to the global update direction.

\bibliography{reference}
\bibliographystyle{elsarticle-num}
\end{document}